\documentclass{article}
\usepackage{iclr2027_conference,times}

\usepackage{amsmath,amsfonts,bm}

\def\eqref#1{equation~\ref{#1}}

\def\1{\bm{1}}

\DeclareMathAlphabet{\mathsfit}{\encodingdefault}{\sfdefault}{m}{sl}
\SetMathAlphabet{\mathsfit}{bold}{\encodingdefault}{\sfdefault}{bx}{n}

\usepackage{hyperref}
\hypersetup{hidelinks}
\usepackage{url}
\usepackage{graphicx}
\usepackage{booktabs}
\usepackage{array}
\usepackage[table]{xcolor}
\usepackage{wrapfig}
\usepackage{amssymb}
\usepackage{subcaption}
\usepackage{enumitem}
\usepackage{algorithm}
\usepackage{algpseudocode}
\usepackage[most]{tcolorbox}
\usepackage{xcolor}
\usepackage{wrapfig}
\usepackage{booktabs}

\definecolor{ucobblue}{RGB}{236,245,255}
\definecolor{succyellow}{RGB}{255,253,238}
\definecolor{ucobmetric}{RGB}{242,249,238}
\definecolor{groupgray}{RGB}{242,242,242}
\newcolumntype{Y}{>{\columncolor{succyellow}}c}

\newcommand{\stepbox}[1]{%
    \begingroup
    \setlength{\fboxsep}{3pt}%
    \colorbox{gray!15}{\strut\textbf{#1}}%
    \endgroup
}

\title{MDSKILL: Learning to Internalize Skill Memory through Multi-View Distillation for Efficient Dynamic Agent Self-Evolution}

\author{
{\small\bfseries
Dian Jin$^{1,2}$, Jinluan Yang$^{1}$,
Kairong Han$^{1}$, Baohong Li$^{1}$,
Zijing Hu$^{1}$, Yi He$^{1}$, Wenlin Zhong$^{1,2}$,}\\
{\small\bfseries
Yunze Tong$^{1}$, Minqin Zhu$^{1,2}$,
Linchuan Li$^{2}$, Yanan Niu$^{2}$,
Kun Kuang$^{1}$\thanks{Corresponding author.}}\\[0.3em]
{\normalfont
$^{1}$Zhejiang University, $^{2}$Kuaishou Technology}\\[0.2em]
{\normalfont
\href{mailto:dianjin@zju.edu.cn}{%
\textcolor{blue!60!black}{\texttt{dianjin@zju.edu.cn}}}}
}

\iclrfinalcopy

\begin{document}

\maketitle

\fancyhead{}
\renewcommand{\headrulewidth}{0pt}

\begin{abstract}
Skill memories encode past experiences as textual prompts to improve reinforcement learning (RL) for self-evolving agents. Complementing RL, On-Policy Self-Distillation(OPSD) provides dense, fine-grained supervision to internalize knowledge from skill-guided trajectories and accelerate policy learning. However, self-generated memories often contain noisy supervision and functionally entangled, redundant, or conflicting knowledge, making direct distillation difficult. Under limited training budgets, these issues can hinder knowledge internalization and even degrade performance. To address this problem, we propose MDSKILL, a multi-view distillation approach for skill memory internalization. MDSKILL organizes skill memories into three complementary functional views—state grounding, action planning, and execution control—and selects high-quality skills using dynamic metrics to construct view-specific and integrated reasoning branches. OPSD then transfers and consolidates complementary experiential knowledge across these branches into a shared policy, enabling efficient skill internalization and continuous agent self-evolution. We evaluate MDSKILL on three agent benchmarks against RL, skill-augmented, and self-distillation baselines that do not rely on stronger external teacher models. Without additional computational overhead, MDSKILL improves upon SOTA performance by up to 7.1\% and 8.1\% on ALFWorld and WebShop. Extensive ablations and evolution analyses further validate its effectiveness in knowledge internalization and policy improvement. The code is available at the \href{https://anonymous.4open.science/r/MDSKILL-CF53}{https://anonymous.4open.science/r/MDSKILL-CF53}.

\end{abstract}

\section{Introduction}
Agentic reinforcement learning (RL) has demonstrated strong performance on multi-turn interactive tasks \citep{shao2024deepseekmath, yang2025qwen3, dong2026agentic, gigpo}. However, sparse and delayed rewards provide limited supervision for policy improvement. Skill memories offer a promising source of experiential knowledge: they summarize lessons from real interaction trajectories and guide subsequent decisions through textual prompts \citep{shinn2023reflexion, wang2023voyager}. Although external skill memories improve task success, agent self-evolution requires internalizing their experiential knowledge into model parameters to retain and apply acquired capabilities without explicit memory retrieval. On-Policy Self-Distillation (OPSD) distills skill-view trajectories into the skill-free branch, providing dense supervision for skill internalization and policy learning \citep{opsd, he2026self, zhang2026embarrassingly}. Combining OPSD with RL enables more effective agent self-evolution and greater performance gains \citep{SDAR, UCOB}.

OPSD uses high-quality reference responses to provide reliable teacher supervision \citep{opsd}. In online agent self-evolution, however, the teacher branch relies on self-generated skill memories that may contain ineffective, harmful, or hallucinated strategies, potentially causing the skill-view branch to underperform its skill-free counterpart \citep{UCOB}. Our observational experiments further suggest that jointly applying skills with different decision functions entangles their roles, while redundancy and conflicts obscure their individual contributions. Consequently, direct distillation can propagate erroneous behaviors and provide inconsistent supervision for useful knowledge, hindering its internalization into model parameters.


\begin{figure*}[t]
    \centering

    \begin{subfigure}[t]{0.57\textwidth}
        \vspace{0pt}
        \centering
        \begin{minipage}[c][5.2cm][c]{\linewidth}
            \centering
            \includegraphics[
                width=\linewidth,
                height=5.2cm,
                keepaspectratio,
                trim={26mm 10mm 25mm 10mm},
                clip
            ]{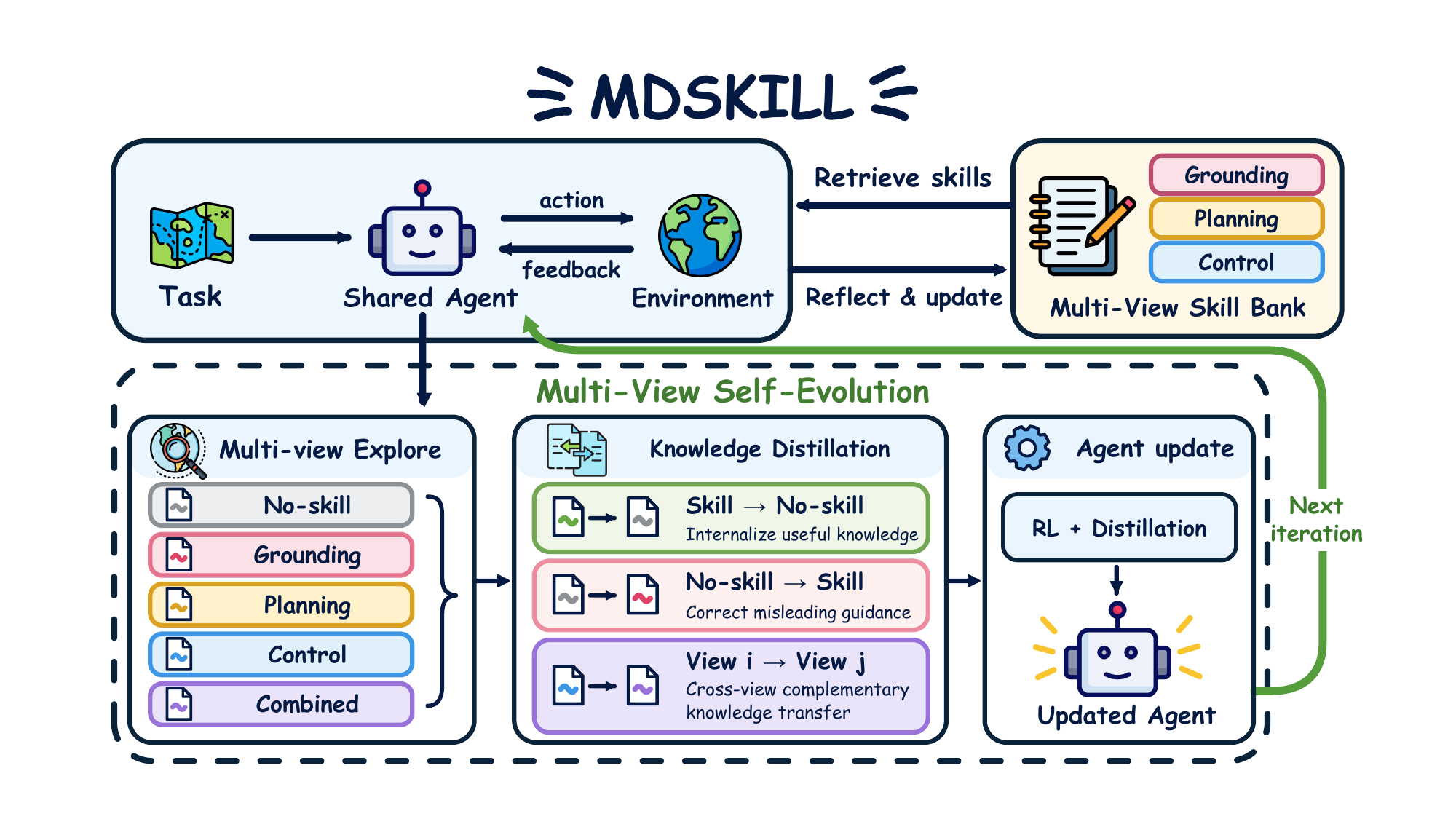}
        \end{minipage}
        \caption{Overall framework of MDSKILL.}
        \label{fig:mdskill_overview}
    \end{subfigure}\hfill%
    \begin{subfigure}[t]{0.43\textwidth}
        \vspace{0pt}
        \centering
        \begin{minipage}[c][5.2cm][c]{\linewidth}
            \centering
            \raisebox{-1mm}[\height][\depth]{%
                \includegraphics[
                    width=\linewidth,
                    height=5.2cm,
                    keepaspectratio
                ]{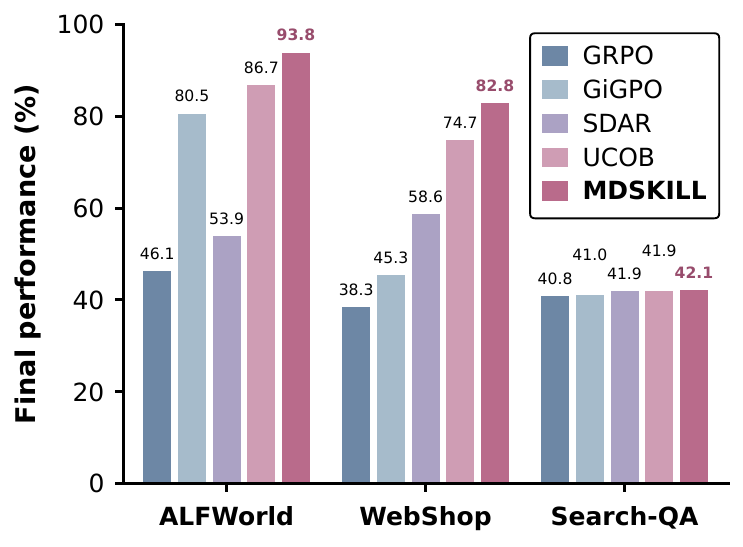}%
            }
        \end{minipage}
        \caption{Main results on Qwen3-1.7B.}
        \label{fig:mdskill_1p7b_results}
    \end{subfigure}

    \par\vspace{5mm}

    \begin{subfigure}[t]{\textwidth}
        \centering
        \includegraphics[
            width=\linewidth,
            height=8.0cm,
            keepaspectratio,
            trim={20mm 13mm 18mm 11mm},
            clip
        ]{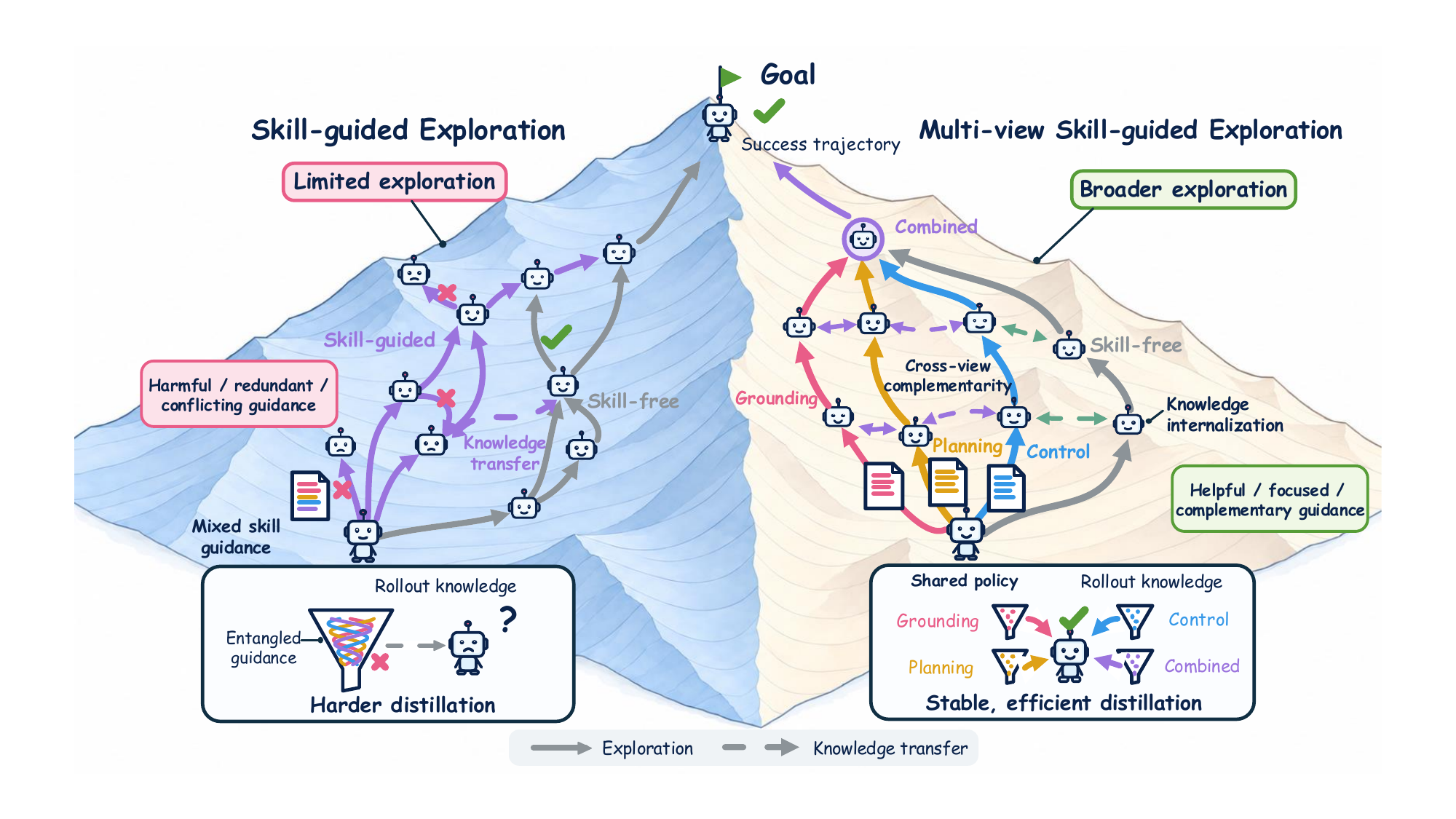}
        \caption{Comparison of multi-view and direct distillation.}
        \label{fig:mdskill_peak}
    \end{subfigure}

    \par\vspace{0mm}
    \caption{Overview, results and intuition on MDSKILL.}
    \label{fig:mdskill_intro}
    \par\vspace{-2mm}
\end{figure*}


Motivated by these observations, we propose MDSKILL, a multi-view distillation framework for skill memory internalization. It uses dynamic value metrics to select skills with additional decision value and organizes them into three complementary functional views: state grounding, action planning, and execution control. These views guide three view-specific branches and an integrated branch alongside a skill-free branch. By disentangling experience according to its decision function, multi-view distillation reduces transfer interference and consolidates complementary knowledge into the shared policy. The skill-view and skill-free branches then co-evolve, generating better trajectories and skill memories that sustain agent self-evolution. Figure~\ref{fig:mdskill_intro} summarizes MDSKILL’s framework, empirical results, and the intuition behind multi-view distillation.

Across three agent benchmarks, MDSKILL outperforms RL, skill-augmented, and self-distillation baselines that do not rely on stronger external teacher models, improving upon SOTA performance by up to 7.1\% and 8.1\% on ALFWorld and WebShop without additional computational overhead. Extensive ablations and skill-evolution analyses support its effectiveness, while gains in both skill-view and skill-free branches indicate successful knowledge internalization rather than continued reliance on external memories. Our main contributions are summarized as follows:
\begin{itemize}[
]
    \item We observe and identify a key challenge in skill-guided agent self-evolution: noisy and functionally entangled skill memories can hinder knowledge internalization and degrade performance.

    \item We propose MDSkill, a multi-view distillation framework that organizes skill memories into complementary decision views and transfers knowledge across skill-view and skill-free branches for more stable and effective experiential knowledge internalization.

    \item Experiments on three agent benchmarks show that MDSKILL improves upon SOTA performance, with extensive ablations and skill-evolution analyses supporting its effectiveness.
\end{itemize}

\section{Preliminaries}

\subsection{Agentic Reinforcement Learning with Skill Memories}

We formulate a multi-turn agent task as a sequential decision-making process. During online training, the agent summarizes experience from successful and failed trajectories as textual \emph{skill memories}, which are stored in an evolving skill bank. Given a task instruction $x$ and interaction history $h_t$, the agent retrieves relevant memories $m_t$ and generates actions through either the skill-free or skill-view branch:
\begin{equation}
\underbrace{a_t \sim \pi_\theta(\cdot \mid x,h_t)}_{\text{skill-free}},
\qquad
\underbrace{a_t \sim \pi_\theta(\cdot \mid x,h_t,m_t)}_{\text{skill-view}}.
\label{eq:agent_policy}
\end{equation}
Both branches share policy parameters $\theta$. Each action yields a new environment observation, producing a trajectory $\tau=(x,o_1,a_1,\ldots,o_T,a_T)$ with return $R(\tau)$. Reinforcement learning optimizes the policy to maximize the expected trajectory return:
\begin{equation}
\mathcal{J}_{\mathrm{RL}}(\theta)
=
\mathbb{E}_{\tau\sim\pi_\theta}[R(\tau)].
\label{eq:rl_objective}
\end{equation}
We adopt GiGPO~\citep{gigpo} as the underlying RL algorithm, combining episode-level and step-level advantages to improve credit assignment under sparse or delayed rewards.

\subsection{On-Policy Self-Distillation}

On-Policy Self-Distillation (OPSD)~\citep{opsd} complements RL with dense token-level supervision. In our setting, the skill-conditioned policy serves as the teacher, transferring knowledge to the skill-free policy through:
\begin{equation}
\mathcal{L}_{\mathrm{OPSD}}
=
\mathbb{E}_{\tau\sim\pi_\theta(\cdot\mid x,m)}
\left[
\frac{1}{|\tau|}
\sum_t
D_{\mathrm{KL}}
\left(
\underbrace{\pi_{\bar{\theta}}(\cdot\mid x,h_t,m_t)}_{\text{teacher}}
\Vert
\underbrace{\pi_\theta(\cdot\mid x,h_t)}_{\text{student}}
\right)
\right].
\label{eq:opsd}
\end{equation}
where $m_t$ denotes retrieved skill memories and $\bar{\theta}$ denotes stop-gradient teacher parameters. This objective directly internalizes skill-guided behavior into the shared policy; we further adapt the transfer direction according to paired trajectory outcomes.

\section{Analysis of Distillation Directions}



We analyze four training variants to examine how distillation direction affects skill-memory internalization: no distillation, forward distillation from the skill-view to the skill-free branch, reverse distillation in the opposite direction, and bidirectional distillation combining both.







\begin{wrapfigure}[16]{r}{0.42\textwidth}
    \centering
    \vspace{-12pt}
    \includegraphics[width=\linewidth]
        {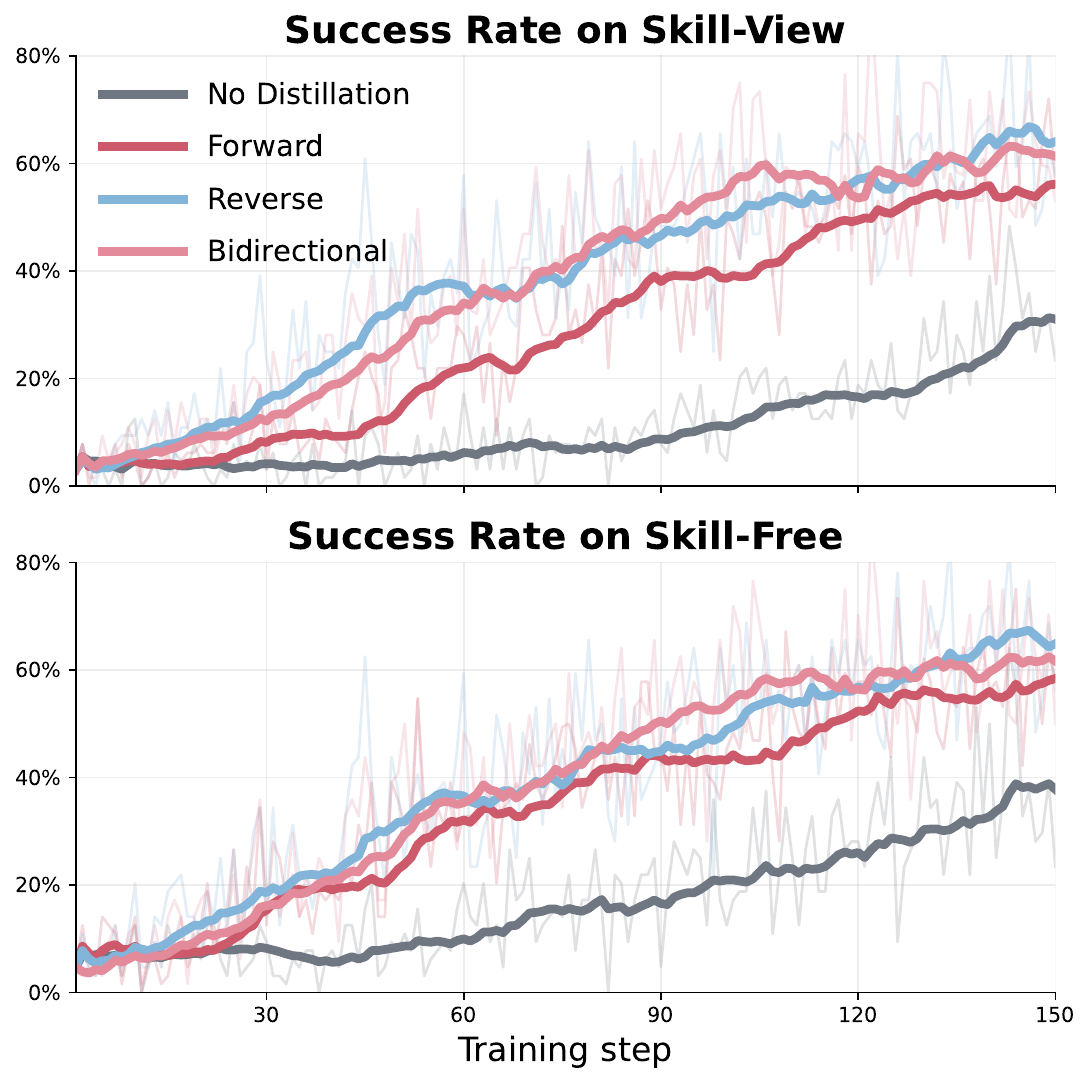}
    \caption{Training success rates under different
        distillation directions on WebShop.}
    \label{fig:distillation_ablations}
\end{wrapfigure}

\paragraph{Distillation effectively internalizes externalized experience.}
As shown in Figure~\ref{fig:distillation_ablations}, all distillation variants outperform no distillation on both branches, demonstrating that distillation effectively converts knowledge externalized in skill memories into the policy’s intrinsic capabilities.

\paragraph{Forward distillation propagates unreliable supervision.}
Forward-only distillation performs substantially worse because the dense supervision signals from the skill-view branch are unreliable. Since both branches share the same policy, this unreliable token-level supervision degrades both skill-view and skill-free behavior.

\begin{wrapfigure}[17]{r}{0.42\textwidth}
    \centering
    \vspace{-20pt}
    \includegraphics[width=\linewidth]
        {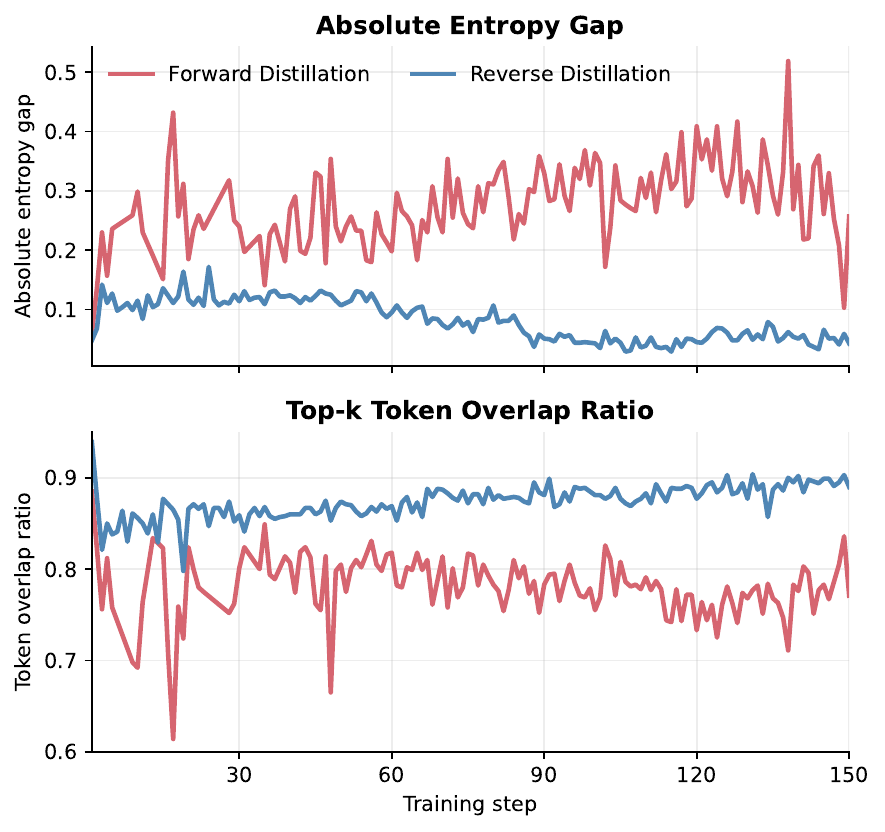}
    \caption{Top-$K$ token overlap ratio and absolute entropy gap under forward and reverse distillation on WebShop.}
    \label{fig:distillation_instability}
\end{wrapfigure}

\paragraph{Reverse-only distillation provides the most effective path to skill internalization.}
Reverse-only distillation achieves the strongest performance on both branches, surpassing even bidirectional distillation. This result highlights distillation’s sensitivity to supervision quality: dense but unreliable signals from forward distillation can dilute or counteract the gains from reverse correction. The advantage of reverse-only distillation thus lies in removing unstable forward supervision, allowing externalized skill knowledge to enter policy parameters indirectly through RL while reverse distillation corrects unreliable skill-guided behavior.

\paragraph{Forward distillation exhibits instability and distributional misalignment.}
To investigate why forward distillation degrades performance, we follow~\citep{li2026rethinking} and analyze Top-$K$ overlap and the absolute entropy gap in Figure~\ref{fig:distillation_instability}. Forward distillation requires the skill-free branch to reproduce skill-conditioned preferences without access to the underlying skill. Its low, volatile overlap and large, fluctuating entropy gap suggest persistent misalignment and unstable supervision, potentially introducing noisy policy updates. In contrast, reverse distillation primarily corrects harmful skill influence and exhibits increasing overlap and a smaller, more stable entropy gap.

\paragraph{Skill noise and knowledge entanglement hinder effective transfer.}
Although forward distillation underperforms reverse distillation, it substantially outperforms no distillation, indicating that skill-guided supervision contains useful knowledge alongside unreliable signals. The challenge therefore lies not simply in a lack of valuable knowledge, but in transferring it effectively. Beyond skill noise, redundant or conflicting guidance from entangled decision functions can obscure useful signals and complicate internalization. These observations motivate dynamic value metrics to prioritize reliable, beneficial skills and multi-view decomposition to separate complementary decision functions, improving supervision quality and reducing transfer interference.
\section{MDSKILL}

\subsection{Overview}


MDSkill is a self-evolution framework that builds on OPSD to aggregate and internalize knowledge from skill memories. It combines dynamic skill selection with multi-view exploration and distillation to internalize complementary knowledge into the shared policy. The improved policy generates better trajectories and higher-quality skill memories, which strengthen policy learning, forming a closed loop of skill evolution, knowledge internalization, and agent self-evolution.

\begin{figure*}[t]
    \centering
    \includegraphics[
        width=\textwidth,
        trim={20mm 10mm 20mm 11mm},
        clip
    ]{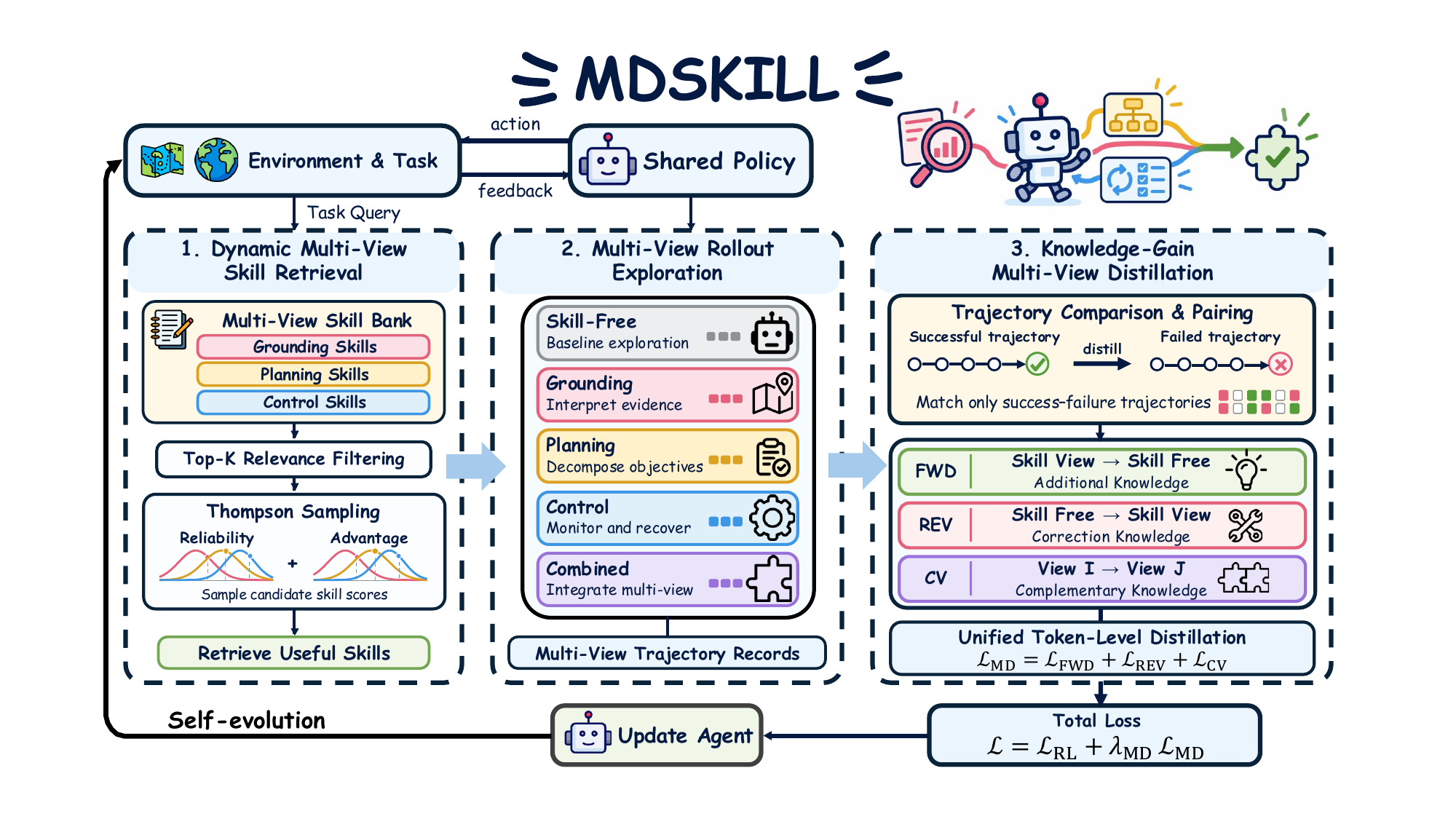}
    \caption{Detailed overview of MDSkill, comprising dynamic skill retrieval, multi-view rollout exploration, and knowledge-gain-driven multi-view distillation.}
    \label{fig:mdskill_overview_main}
\end{figure*}

\subsection{Dynamic Skill Metrics}
\label{sec:dynamic_skill_value}

Skill value estimation identifies useful memories and filters harmful ones, but their value changes as the policy evolves: previously useful skills may become redundant through internalization or be superseded by better strategies. MDSKILL therefore maintains two dynamic, policy-relative estimates, \emph{Reliability} and \emph{Advantage}, updated with new interaction outcomes throughout training. Temporal decay gives recent evidence greater weight, preventing historical performance from dominating the assessment of a skill’s current value.

\paragraph{Dynamic Maintenance.}
For a rollout using skill $m$, let $y\in\{0,1\}$ denote task success and $\bar{y}_{\mathrm{free}}$ the average success of skill-free rollouts on the same task. The corresponding evidence is $z^{\mathrm{rel}}=y$ and $z^{\mathrm{adv}}=\frac{1+y-\bar{y}_{\mathrm{free}}}{2}$. For each estimate $d\in\{\mathrm{rel},\mathrm{adv}\}$, we maintain a Beta distribution with a uniform prior and temporally discounted evidence:
\begin{equation}
w_{t,i}=2^{-\frac{t-t_i}{H}},
\qquad
\alpha^d_{m,t}=1+\sum_{i\in\mathcal{E}_m(t)}w_{t,i}z_i^d,
\qquad
\beta^d_{m,t}=1+\sum_{i\in\mathcal{E}_m(t)}w_{t,i}(1-z_i^d).
\label{eq:skill_evidence}
\end{equation}
where $\mathcal{E}_m(t)$ denotes the observations associated with skill $m$, $t_i$ is the observation step, and $H$ is the evidence half-life. The decay weight $w_{t,i}$ emphasizes recent outcomes and discounts older evidence. More effective evidence concentrates the Beta distributions, making well-supported skill values easier to distinguish while preserving uncertainty for less-evaluated skills.

\paragraph{Thompson-Sampled Retrieval.}
We first retain the Top-$K$ skills with the highest cosine similarity between the embeddings of the current query and each skill's associated context. Within this candidate set, we apply Thompson sampling to rank skills by their sampled \emph{Reliability} and \emph{Advantage}:
\begin{equation}
\mathcal{C}_t(x)=\operatorname{TopK}_{m\in\mathcal{M}_t}\operatorname{sim}(x,c_m),
\qquad
\tilde{q}^{d}_{m,t}\sim\operatorname{Beta}(\alpha^d_{m,t},\beta^d_{m,t}),
\qquad
S_t(m)=\tilde{q}^{\mathrm{rel}}_{m,t}+\tilde{q}^{\mathrm{adv}}_{m,t}.
\label{eq:dynamic_skill_score}
\end{equation}
where $d\in\{\mathrm{rel},\mathrm{adv}\}$, $c_m$ denotes the retrieval context associated with skill $m$, and $\mathcal{M}_t$ is the eligible skill bank. We first select the top-$K$ most relevant skills from $\mathcal{C}_t(x)$ and apply Thompson sampling to their value metrics. Both sampled metrics lie in $[0,1]$, providing a common scale for aggregation. This balances exploitation of well-supported skills with exploration of uncertain candidates, reducing the risk that accidental successes cause particular skills to dominate retrieval.


\subsection{Multi-View Skill Rollouts}
\label{sec:multi_view_skill}

High-value experience can support different aspects of a decision. MDSkill organizes skill memory around three complementary functions---state grounding, action planning, and execution control---and uses these views to guide exploration and selective knowledge transfer into a shared policy.

\subsubsection{Functional Skill Views}

We generate skills through view-specific reflection prompts, each directing the agent to extract a distinct type of reusable knowledge from its interaction experience. The prompts are adapted to each environment, and the resulting skills are maintained in three functional pools, $\mathcal{M}_{\mathrm{g}}$, $\mathcal{M}_{\mathrm{p}}$,  $\mathcal{M}_{\mathrm{c}}$.

\paragraph{State grounding.}
This view addresses \emph{what is known about the current situation}. Grounding skills identify task-relevant observations, object states, constraints, and missing information. They guide the agent to inspect its environment, resolve uncertainty, and distinguish verified facts from assumptions before committing to an action. Exploration under this view emphasizes obtaining and interpreting evidence that makes subsequent decisions better informed.

\begin{wrapfigure}[23]{r}{0.45\textwidth}
    \centering
    \vspace{-10pt}
    \includegraphics[
        width=\linewidth,
        trim=108mm 5mm 108mm 2mm,
        clip
    ]{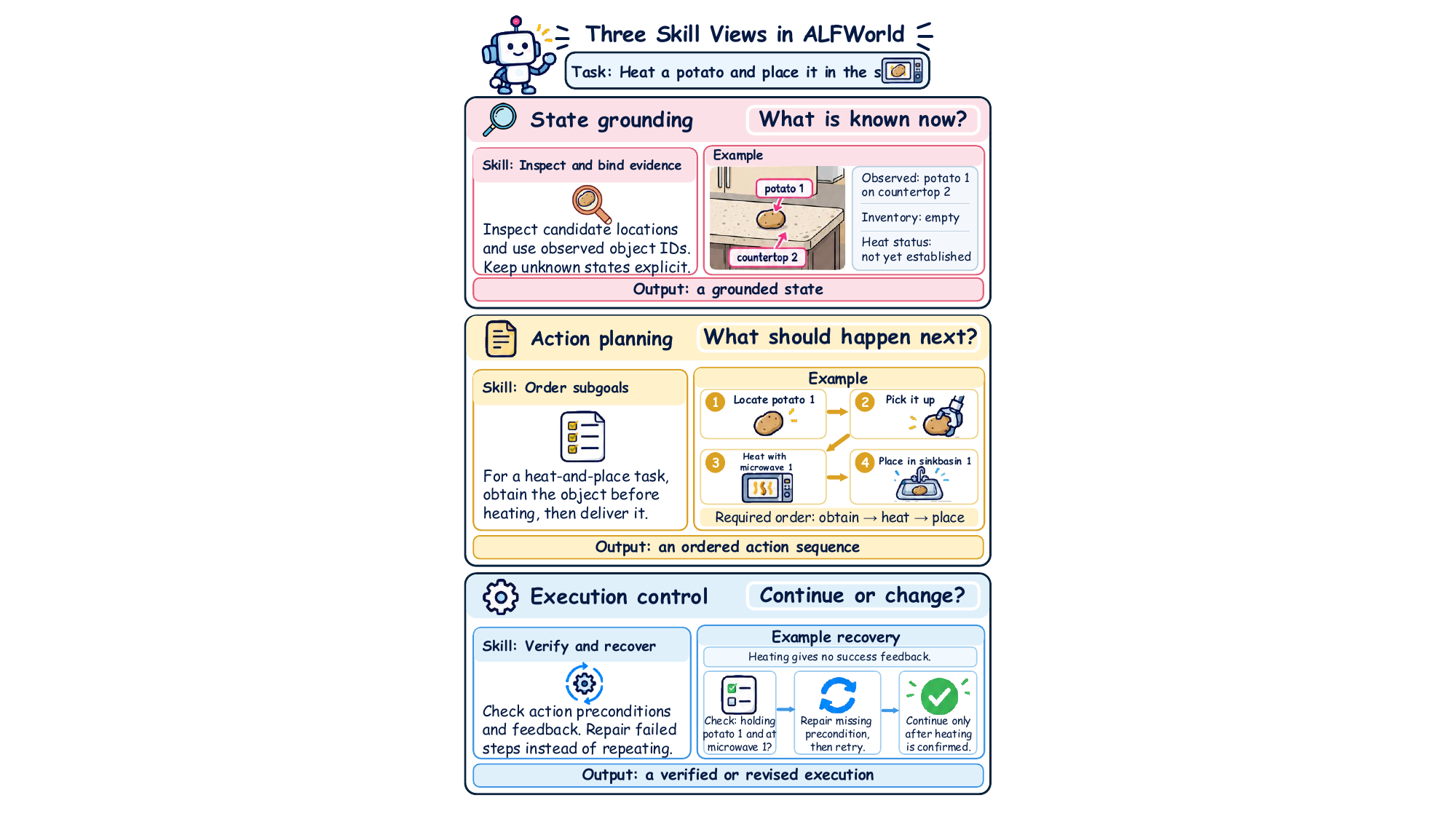}
    \caption{An example of multi-view skill guidance in ALFWorld.}
    \label{fig:potato2}
\end{wrapfigure}

\paragraph{Action planning.}
This view addresses \emph{what to do next given the information}.
Planning skills describe how to select actions, establish subgoals,
and order operations toward task completion. They direct exploration
toward alternative action sequences and reusable strategies for
making progress.

\paragraph{Execution control.}
This view addresses \emph{whether ongoing execution should continue or change}. Control skills focus on checking action preconditions, detecting ineffective behavior, verifying intermediate outcomes, and recovering from errors. They guide the agent to reconsider a plan when observations contradict its expectations, rather than repeatedly following an unsuccessful course of action.


\paragraph{Combined branch.}
This branch integrates knowledge from three skill views, allowing
grounding, planning, and control to complement one another. It connects
observation interpretation with action selection and execution
monitoring, enabling the agent to adjust its plan as new evidence
emerges. The resulting trajectories capture interactions among
decision functions and provide an integrated path alongside the
view-specific branches. Implementation details for functional skills
are provided in Appendix~\ref{sec:functional_skill_reflection}.

\subsubsection{Multi-View Distillation}

\paragraph{Knowledge-Gain Gate.}
Multi-view rollouts provide diverse experiential knowledge, while
distillation requires reliable fine-grained supervision for effective
transfer. For precise distillation matching, following
UCOB~\citep{UCOB}, we group decision records that share the same task
and anchor state. Within each anchor group $a$, let
$\mathcal{T}_a^{0}$ denote the skill-free trajectories and
$\mathcal{T}_a^{+}
=\bigcup_{v\in\mathcal{V}}\mathcal{T}_a^{v}$ denote the skill-view
trajectories, where $\mathcal{V}$ contains the grounding, planning,
control, and combined views. Given the trajectory-level success
indicator $y(\tau)\in\{0,1\}$, we define the numbers of successful
skill-free and skill-view trajectories as
\begin{equation}
n_a^{0}
=
\sum_{\tau\in\mathcal{T}_a^{0}} y(\tau),
\qquad
n_a^{+}
=
\sum_{\tau\in\mathcal{T}_a^{+}} y(\tau).
\label{eq:anchor_success_counts}
\end{equation}
The strict and relaxed knowledge-gain gates are defined as
\begin{equation}
\gamma_a^{\mathrm{strict}}
=
\mathbb{I}\!\left[
n_a^{+}>n_a^{0}
\land n_a^{0}=0
\right],
\qquad
\gamma_a^{\mathrm{relaxed}}
=
\mathbb{I}\!\left[n_a^{+}>n_a^{0}\right].
\label{eq:distillation_gates}
\end{equation}

\paragraph{Teacher-student Grouping.}
Teacher-student grouping for distillation depends on trajectory outcomes and the knowledge-gain gate. For each teacher-student pair $(\tau_i,\tau_j)$ in anchor $a$, we define
\begin{equation}
g_a(\tau_t,\tau_s)
=
\mathbb{I}[y(\tau_t)>y(\tau_s)] \cdot
\begin{cases}
\gamma_a,
& v_t\neq 0,\ v_s=0 \quad \text{(forward)},\\[3pt]
1-\gamma_a,
& v_t=0,\ v_s\neq 0 \quad \text{(reverse)},\\[3pt]
\gamma_a,
& v_t\neq 0,\ v_s\neq 0,\ v_t\neq v_s
\quad \text{(cross-view)},\\[3pt]
0,
& \text{otherwise}.
\end{cases}
\label{eq:pairwise_distillation_gate}
\end{equation}
where $v_i=0$ denotes the skill-free branch and
$g_a(\tau_t,\tau_s)=1$ identifies a valid distillation pair.

Distillation takes three forms: (1) \textbf{Forward distillation} transfers successful skill-view behavior to failed skill-free trajectories when the knowledge-gain gate is satisfied, enabling the policy to internalize useful skill-guided behavior; (2) \textbf{Reverse distillation} uses successful skill-free behavior to correct failed skill-guided trajectories when the policy can already solve the task without skills; (3)\textbf{Cross-view distillation} transfers knowledge from a successful skill view to unsuccessful alternative views, enabling complementary knowledge exchange across decision functions. Strict and relaxed variants of the distillation gate are discussed in
Appendix~\ref{sec:distillation_gate_discussion}.

\paragraph{Multi-View Distillation Objective.}
All forward, reverse, and cross-view pairs share a unified
token-level objective. For each valid pair $(\tau_t,\tau_s)$,
the teacher response is evaluated under both contexts.
We collect valid token instances and define the joint loss as
\begin{equation}
\begin{aligned}
\mathcal{I}_{\mathrm{MD}}
&=
\{(a,t,s,\ell):
g_a(\tau_t,\tau_s)=1,\ 1\leq\ell\leq N_{t,s}\},\\[4pt]
\mathcal{L}_{\mathrm{MD}}
&=
\frac{1}{|\mathcal{I}_{\mathrm{MD}}|}
\sum_{(a,t,s,\ell)\in\mathcal{I}_{\mathrm{MD}}}
D_{\mathrm{KL}}\!\left(
\operatorname{sg}[\bar{p}_{t,\ell}]
\Vert
\bar{p}_{s,\ell}
\right).
\end{aligned}
\label{eq:multiview_distillation_objective}
\end{equation}
where $N_{t,s}$ denotes the valid response token count,
and $\bar{p}_{t,\ell}$ and $\bar{p}_{s,\ell}$ are normalized
distributions over the teacher's Top-$K_{\mathrm{D}}$ token support \citep{fu2026revisiting, UCOB}. The operator $\operatorname{sg}[\cdot]$ stops teacher gradients.
The complete training objective is
\begin{equation}
\mathcal{L}
=
\mathcal{L}_{\mathrm{RL}}
+
\lambda_{\mathrm{MD}}\mathcal{L}_{\mathrm{MD}}
+
\beta_{\mathrm{KL}}\mathcal{L}_{\mathrm{KL}}.
\label{eq:mdskill_objective}
\end{equation}
where $\lambda_{\mathrm{MD}}$ controls the distillation weight,
$\mathcal{L}_{\mathrm{KL}}$ regularizes policy divergence from
the reference policy, and $\beta_{\mathrm{KL}}$ controls its strength.

\section{Experiments}
\label{sec:experiments}

\subsection{Experimental Setup}
\label{sec:experimental_setup}

\paragraph{Benchmarks}
Following the evaluation suite of SDAR~\citep{SDAR}, we evaluate on three agent benchmarks. \textsc{ALFWorld}~\citep{shridhar2020alfworld} tests household navigation and object manipulation across six task families: Pick, Look, Clean, Heat, Cool, and Pick2. \textsc{WebShop}~\citep{yao2022webshop} requires agents to find and purchase products satisfying natural-language requests; we use the established 128-task evaluation split. \textsc{Search-QA}~\citep{jin2025search} evaluates search-augmented question answering across three single-hop and four multi-hop datasets. We report success rates on ALFWorld, task scores and success rates on WebShop, and answer accuracy on Search-QA.

\paragraph{Models and baselines.}
We use Qwen2.5-7B-Instruct, Qwen2.5-3B-Instruct~\citep{yang2025qwen3}, and Qwen3-1.7B~\citep{yang2025qwen3} as backbones. We compare against three baseline groups: (1) skill-free RL methods, GRPO~\citep{shao2024deepseekmath} and GiGPO~\citep{gigpo}; (2) skill-augmented methods, SkillRL~\citep{xia2026skillrl}, D2Skill~\citep{tu2026dynamic}, SkillC~\citep{lin2026skillc}, SAPO~\citep{zhang2026co}, and Skill1~\citep{shi2026skill1}; and (3) self-distillation methods, OPSD~\citep{opsd}, RLSD~\citep{yang2026self}, Skill-SD~\citep{wang2026skill}, SDAR~\citep{SDAR}, StepOPSD~\citep{zhang2026stepopsd} and UCOB~\citep{UCOB}. These baselines provide broad coverage of agent self-evolution methods. For fair comparison, MDSKILL follows the same experimental settings as the corresponding baselines; details are provided in Appendix~\ref{implementation_details}.

\definecolor{backbonegray}{RGB}{226,226,226}
\definecolor{metricgray}{RGB}{243,243,243}
\definecolor{mdskillred}{RGB}{252,237,237}
\definecolor{referenceblue}{RGB}{230,230,255}
\begin{table*}[t]
    \centering
    \caption{Main results on \textsc{ALFWorld}
    (six tasks and overall success rates),
    \textsc{WebShop} (score/success rate), and \textsc{Search-QA}
    (seven datasets and average accuracy).
    Best and second-best results within each backbone are
    \textbf{bolded} and \underline{underlined}, respectively.
    $\dagger$ denotes results from original papers;
    $^{*}$ denotes results reported by SDAR and UCOB.}
    \label{tab:main_results}

    \scriptsize
    \setlength{\tabcolsep}{1.5pt}
    \renewcommand{\arraystretch}{1.00}

    \resizebox{\textwidth}{!}{%
    \begin{tabular}{l|cccccc>{\columncolor{metricgray}}c
                    !{\hspace{1pt}}|!{\hspace{1pt}}
                    c>{\columncolor{metricgray}}c
                    !{\hspace{1pt}}|!{\hspace{1pt}}
                    ccccccc>{\columncolor{metricgray}}c}
        \toprule
        &
        \multicolumn{7}{c}{\textbf{\textsc{ALFWorld}}} &
        \multicolumn{2}{c}{\textbf{\textsc{WebShop}}} &
        \multicolumn{8}{c}{\textbf{\textsc{Search-QA}}}
        \\
        \cmidrule(lr){2-8}
        \cmidrule(lr){9-10}
        \cmidrule(lr){11-18}

        \textbf{Method}
        & \textbf{Pick} & \textbf{Look} & \textbf{Clean}
        & \textbf{Heat} & \textbf{Cool} & \textbf{Pick2}
        & \textbf{Succ} & \textbf{Score} & \textbf{Succ}
        & \textbf{NQ} & \textbf{Triv} & \textbf{Pop}
        & \textbf{Hotp} & \textbf{2Wk} & \textbf{MuS}
        & \textbf{Bam} & \textbf{Avg}
        \\
        \midrule

        \rowcolor{metricgray}
        \multicolumn{18}{c}{\textbf{Qwen3-1.7B}}
        \\
        \midrule

        GRPO$^{*}$
        & 71.1 & 41.7 & 36.4 & 40.0 & 31.8 & 31.6
        & 46.1 & 67.3 & 38.3
        & 40.0 & 58.9 & 43.5 & 35.4 & 30.3
        & \underline{12.0} & \textbf{65.7} & 40.8
        \\

        GiGPO
        & 90.3 & 61.5 & \underline{92.3} & 71.4 & 69.2 & \underline{79.2}
        & 80.5 & 70.0 & 45.3
        & 41.1 & 59.3 & 45.1 & 34.1 & 32.7
        & 10.5 & 64.5 & 41.0
        \\

        Skill1$^{*}$
        & 71.4 & 54.5 & 59.4 & 72.7 & 60.0 & 73.7
        & 65.6 & 83.2 & 61.7
        & \underline{42.7} & \textbf{59.8} & \textbf{46.0}
        & 35.4 & \underline{34.8} & \underline{12.0}
        & \textbf{65.7} & \textbf{42.3}
        \\

        RLSD$^{*}$
        & 50.0 & 37.5 & 61.5 & 19.0 & 50.0 & 21.1
        & 42.2 & 74.0 & 50.8
        & 38.6 & 57.3 & 43.0 & 34.5 & 34.1
        & 11.5 & \underline{65.3} & 40.6
        \\

        Skill-SD$^{*}$
        & 52.9 & 37.5 & 69.2 & 42.9 & 60.0 & 36.8
        & 52.3 & 81.8 & 53.9
        & 39.1 & 57.5 & 45.4 & 34.8 & 34.1
        & 10.7 & 64.1 & 40.8
        \\

        SDAR$^{*}$
        & 73.5 & 25.0 & 76.9 & 33.3 & 40.0 & 36.8
        & 53.9 & 76.8 & 58.6
        & 39.7 & 58.9 & 45.3 & 35.9 & \textbf{35.5}
        & \textbf{12.6} & \underline{65.3} & 41.9
        \\

        StepOPSD$^{\dagger}$
        & 64.7 & 44.4 & 56.5 & 60.9 & 42.1 & 55.0
        & 56.3 & -- & --
        & 40.5 & 59.4 & 44.4 & 37.1 & 32.0
        & 11.6 & 64.1 & 41.3
        \\

        UCOB
        & \underline{93.1} & \underline{76.9} & \textbf{100}
        & \underline{84.2} & \textbf{81.8} & 76.2
        & \underline{86.7} & \textbf{90.3} & \underline{74.7}
        & \textbf{43.6} & 58.7 & \underline{45.5}
        & \underline{37.8} & 32.3 & 11.2 & 64.1
        & 41.9
        \\

        \rowcolor{mdskillred}
        \textbf{\textsc{MDSkill}}
        & \textbf{97.4} & \textbf{92.9} & \textbf{100}
        & \textbf{100} & \underline{81.0} & \textbf{88.9}
        & \textbf{93.8} & \underline{88.7} & \textbf{82.8}
        & 41.7 & \underline{59.6} & 45.1
        & \textbf{38.0} & 33.8 & \underline{12.0}
        & 64.8 & \underline{42.1}
        \\



        \midrule
        \rowcolor{metricgray}
        \multicolumn{18}{c}{\textbf{Qwen2.5-3B-Instruct}}
        \\
        \midrule

        GRPO$^{*}$
        & 91.2 & 62.5 & 96.2 & 61.9 & 65.0 & 47.4
        & 75.0 & 79.8 & 63.3
        & 39.3 & 60.6 & 41.1 & 37.4 & 34.6
        & 15.4 & 26.4 & 36.4
        \\

        GiGPO
        & 89.7 & 76.9 & \textbf{100} & 78.9 & 81.8 & 71.4
        & 84.4 & 83.8 & 70.3
        & 43.0 & 60.5 & \textbf{46.9} & 36.2 & 35.5
        & 12.1 & 64.1 & 42.6
        \\

        Skill1$^{*}$
        & 93.8 & 58.3 & \underline{96.4} & \underline{92.9}
        & \textbf{86.4} & \textbf{95.0}
        & 89.1 & 84.1 & 75.0
        & 44.2 & 60.3 & 44.8 & 39.2 & 35.9
        & 13.1 & 64.9 & 43.2
        \\

        RLSD$^{*}$
        & 87.9 & 75.0 & 90.9 & 75.0 & 73.1 & 68.4
        & 79.7 & 84.4 & 66.4
        & 41.5 & 58.6 & 42.3 & \textbf{40.4} & \underline{40.2}
        & \textbf{16.8} & \textbf{66.9} & 43.8
        \\

        Skill-SD$^{*}$
        & 88.2 & 50.0 & 96.2 & 52.4 & 65.0 & 57.9
        & 73.4 & 75.9 & 64.0
        & 44.4 & 60.4 & 44.0 & \underline{39.5} & \textbf{40.4}
        & 15.4 & 64.9 & \underline{44.1}
        \\

        SDAR$^{*}$
        & \underline{97.1} & 62.5 & \textbf{100} & 61.9 & 75.0 & 84.2
        & 84.4 & 85.0 & 68.0
        & \underline{44.8} & 58.1 & 44.3 & 38.6 & 36.2
        & \underline{15.7} & 66.1 & 43.4
        \\

        StepOPSD$^{\dagger}$
        & \underline{97.1} & 66.7 & 87.0 & 79.1 & 78.9 & \textbf{95.0}
        & 83.6 & -- & --
        & \textbf{45.0} & \textbf{61.6} & \underline{46.2}
        & \underline{39.5} & 39.5 & 14.4 & 65.3
        & \textbf{44.5}
        \\

        UCOB
        & \textbf{100} & \underline{90.0} & \textbf{100}
        & 88.9 & 76.0 & 88.2
        & \underline{92.2} & \textbf{86.1} & \underline{77.3}
        & 44.7 & \underline{61.4} & \underline{46.2}
        & 37.5 & 36.2 & 12.2 & \underline{66.3} & 43.5
        \\

        \rowcolor{mdskillred}
        \textbf{\textsc{MDSkill}}
        & \textbf{100} & \textbf{100} & 96.3
        & \textbf{93.3} & \underline{85.0} & \underline{94.1}
        & \textbf{95.3} & \underline{85.6} & \textbf{80.5}
        & 44.2 & 58.9 & 45.6 & 38.9 & 37.5
        & 14.2 & 64.3 & 43.4
        \\

        \midrule
        \rowcolor{metricgray}
        \multicolumn{18}{c}{\textbf{Qwen2.5-7B-Instruct}}
        \\
        \midrule

        GRPO$^{*}$
        & 91.2 & 87.5 & 96.2 & 81.0 & 65.0 & 57.9
        & 81.2 & 80.9 & 72.6
        & 45.1 & 63.7 & 44.0 & 43.6 & 43.2
        & 16.8 & 37.6 & 42.0
        \\

        GiGPO$^{\dagger}$
        & 97.7 & 82.7 & \underline{98.8} & 83.7 & 89.3 & 79.2
        & 90.8 & 84.4 & 72.8
        & 46.4 & \underline{64.7} & 46.1 & 41.6 & 43.6
        & 18.9 & 68.9 & 47.2
        \\

        SkillRL$^{\dagger}$
        & 97.9 & 71.4 & 90.0 & 90.0 & \underline{95.5} & 87.5
        & 89.9 & 85.2 & 72.7
        & 45.9 & 63.3 & 45.9 & 43.2 & 40.3
        & 20.2 & \textbf{73.8} & 47.5
        \\

        D2Skill$^{\dagger}$
        & 93.8 & 77.8 & 94.7 & 95.0 & \underline{95.5} & 72.0
        & 87.8 & \underline{91.1} & 80.5
        & \textbf{48.7} & 63.4 & 44.9 & \textbf{47.5} & 43.7
        & \underline{21.0} & 67.7 & 48.1
        \\

        SkillC$^{\dagger}$
        & 88.5 & 78.6 & 91.2 & 94.7 & \textbf{100.0} & 88.9
        & 90.6 & 85.6 & 74.0
        & -- & -- & -- & -- & -- & -- & -- & --
        \\

        SAPO$^{\dagger}$
        & \underline{98.7} & 73.9 & 98.1 & 92.6 & 85.0 & \underline{89.2}
        & 92.2 & 90.5 & 78.1
        & \underline{48.4} & 62.9 & 46.7 & 45.0 & 45.2
        & 18.3 & 46.4 & 44.7
        \\

        Skill1$^{*}$
        & \textbf{100.0} & 83.3 & 89.5 & \textbf{100.0} & 88.5 & 85.0
        & 92.2 & 86.2 & 77.3
        & 46.8 & 46.3 & \underline{47.8} & 43.7 & 39.3
        & 18.2 & 70.6 & 44.7
        \\

        RLSD$^{*}$
        & \textbf{100.0} & 87.5 & 92.3 & 58.8 & 80.0 & 65.2
        & 82.0 & 87.4 & 77.3
        & 46.8 & 63.0 & 44.4 & 45.5 & \textbf{48.9}
        & \textbf{21.5} & \underline{73.0} & \textbf{49.0}
        \\

        Skill-SD$^{*}$
        & 93.9 & \underline{93.8} & 90.9 & \textbf{100.0} & 69.2 & 68.4
        & 85.1 & 86.1 & 76.5
        & 47.1 & 64.5 & \underline{47.8} & 44.2 & 42.1
        & 20.2 & 69.0 & 47.8
        \\

        SDAR$^{*}$
        & 94.7 & 75.0 & \textbf{100.0} & 86.7 & 68.2 & 78.9
        & 85.9 & 89.4 & \underline{82.8}
        & 46.3 & 63.5 & \textbf{48.2} & 43.8 & \underline{48.4}
        & 19.6 & \underline{73.0} & \textbf{49.0}
        \\

        UCOB
        & \textbf{100} & 81.8 & 96.2 & \textbf{100} & 91.3 & 85.7
        & \underline{93.8} & \textbf{91.3} & 80.5
        & 46.8 & 64.2 & 47.0 & \underline{46.1} & 42.2
        & 18.7 & 71.0 & 48.0
        \\

        \rowcolor{mdskillred}
        \textbf{\textsc{MDSkill}}
        & 97.0 & \textbf{100} & \textbf{100} & \underline{95.2}
        & 91.3 & \textbf{94.4}
        & \textbf{96.1} & 88.4 & \textbf{84.4}
        & 46.5 & \textbf{65.3} & 46.8 & 45.9 & 44.3
        & 18.4 & 71.7 & \underline{48.4}
        \\

        \bottomrule
    \end{tabular}%
    }

    \vspace{-2em}
\end{table*}

\subsection{Main Results}
\label{sec:main_results}

\paragraph{Overall Performance.}
Table~\ref{tab:main_results} shows that MDSkill achieves the best overall success rates on both ALFWorld and WebShop across all three backbones. On ALFWorld, MDSkill outperforms the strongest baseline UCOB by 7.1\%, 3.1\%, and 2.3\% with Qwen3-1.7B, Qwen2.5-3B, and Qwen2.5-7B, respectively. On WebShop, it improves the success rate by 8.1\%, 3.2\%, and 3.9\%, while maintaining competitive task scores. MDSKILL also achieves competitive performance on Search-QA across model scales. Compared with interactive decision-making tasks, Search-QA depends more on retrieval quality, evidence comprehension, and answer synthesis, and less on recurring environment states where reusable skills offer consistent guidance.

\paragraph{Knowledge Internalization.}
The consistent gains on ALFWorld and WebShop demonstrate the value of internalizing skill memories from complementary decision perspectives. The gains are particularly pronounced with the smaller Qwen3-1.7B model which has fewer task-relevant priors and reusable interaction patterns, with success rates increasing from 86.7\% to 93.8\% on ALFWorld and from 74.7\% to 82.8\% on WebShop. By organizing skill memories into complementary views and distilling successful behaviors into the shared policy, MDSkill mitigates the knowledge disadvantage of smaller models, while the improved policy produces stronger trajectories and higher-quality skills that further reinforce knowledge internalization and agent self-evolution.

\begin{figure*}[t]
    \centering
    \captionsetup[subfigure]{skip=4pt}

    \begin{subfigure}[b]{0.6\textwidth}
        \centering
        \includegraphics[width=\linewidth]
            {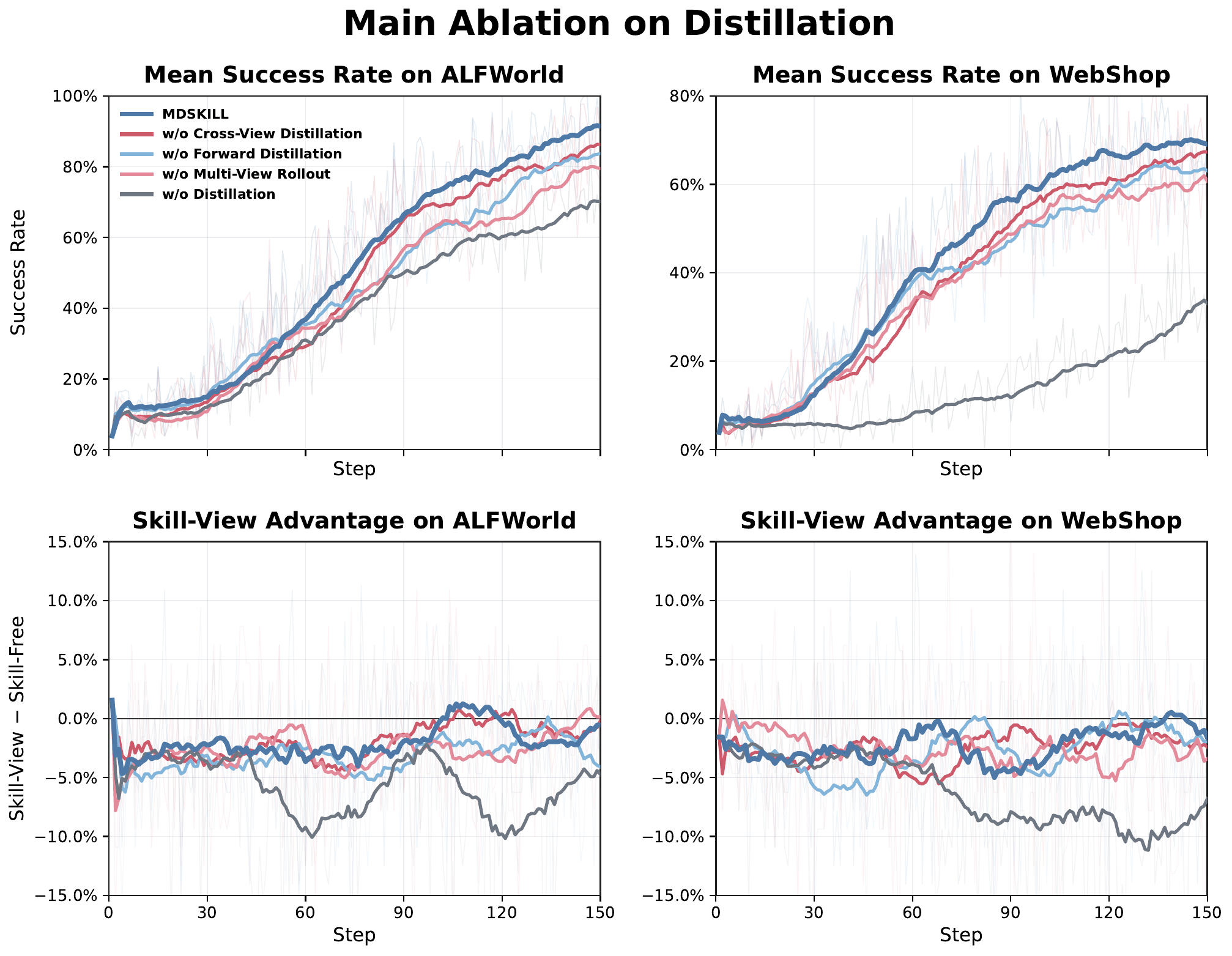}
        \caption{Main ablation across training.}
        \label{fig:main_ablation}
    \end{subfigure}\hfill%
    \begin{minipage}[b]{0.4\textwidth}
        \centering

        \begin{subfigure}[b]{\linewidth}
            \centering
            \includegraphics[width=\linewidth]
                {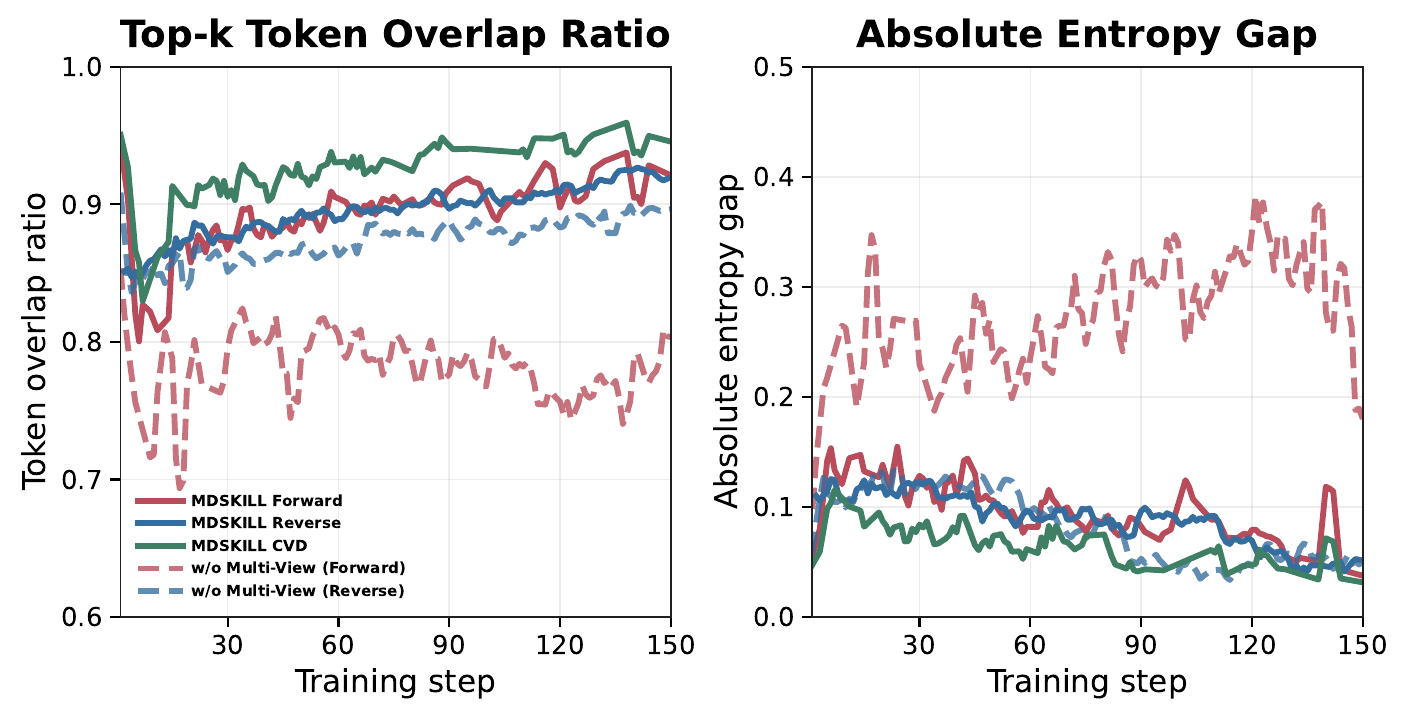}
            \caption{Distributional stability of distillation.}
            \label{fig:distillation_stability}
        \end{subfigure}

        \par\vspace{0.6em}

        \begin{subfigure}[b]{\linewidth}
            \centering
            \raisebox{2.3mm}[\height][\depth]{%
                \includegraphics[width=\linewidth]
                    {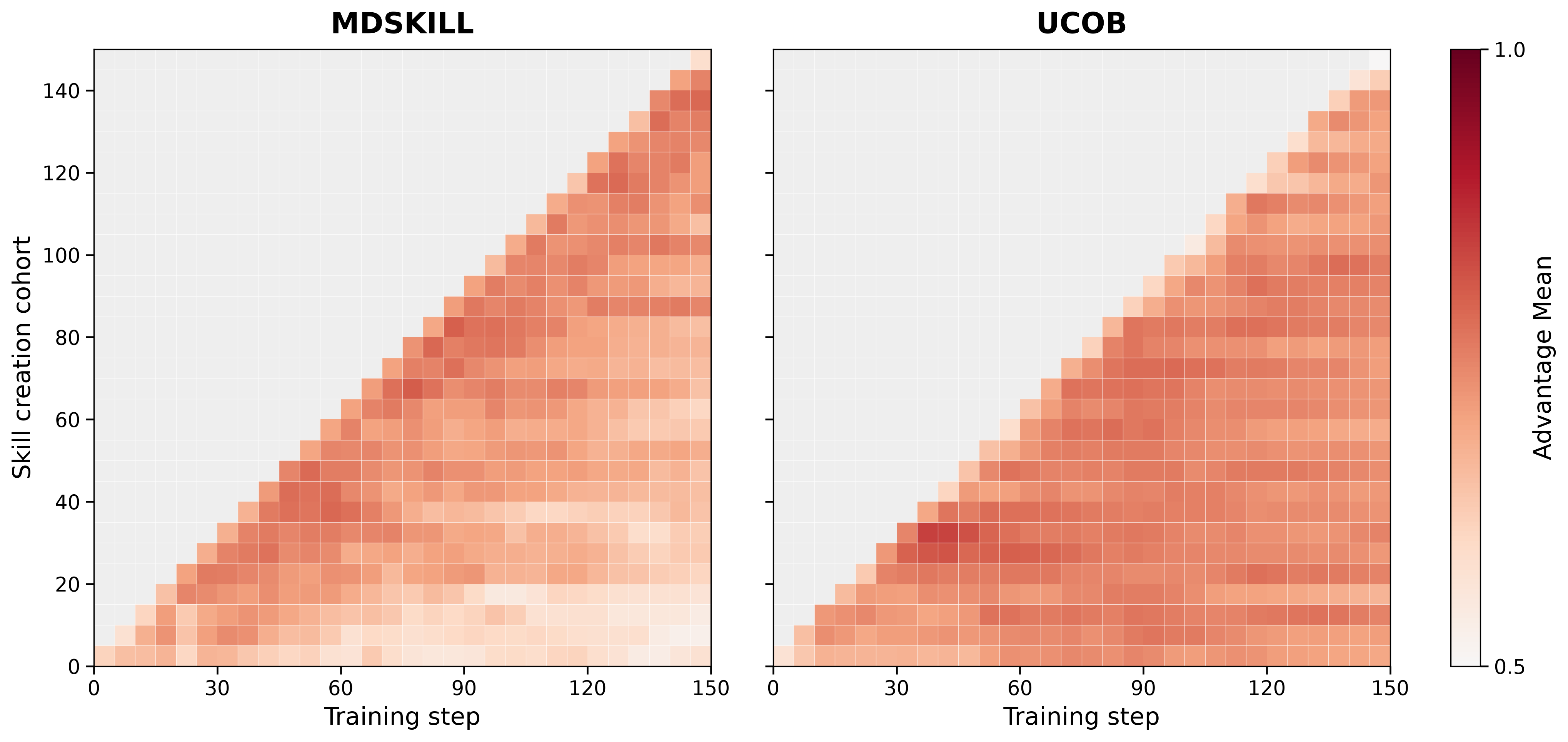}%
            }
            \caption{Dynamic evolution of retrieved skills.}
            \label{fig:skill_evolution}
        \end{subfigure}
    \end{minipage}

    \par\vspace{0.8em}

    \begin{subfigure}[t]{\textwidth}
        \centering
        \includegraphics[
            width=\linewidth,
            trim={0mm 0mm 0mm 15mm},
            clip
        ]{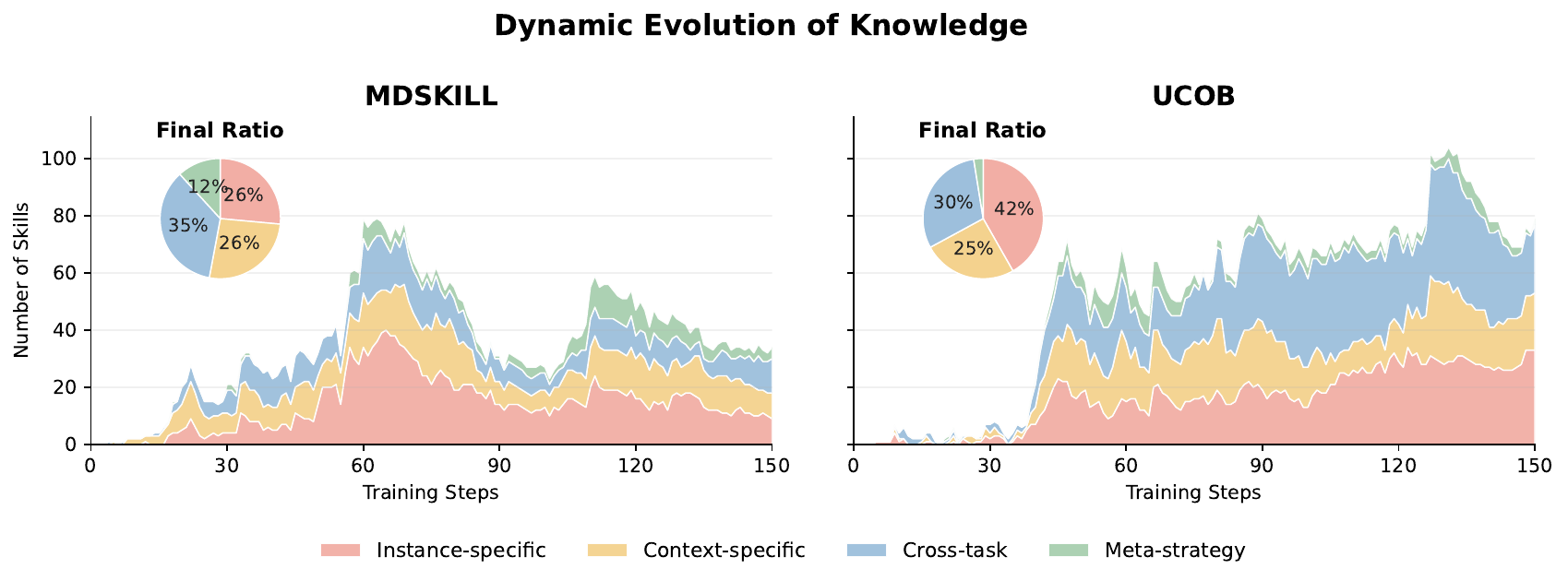}
        \caption{Dynamic evolution of skill knowledge composition.}
        \label{fig:knowledge_evolution}
    \end{subfigure}

    \caption{Ablation study of MDSKILL.}
    \label{fig:mdskill_ablation_analysis}
    \vspace{-2em}
\end{figure*}

\subsection{Ablation Studies}
\label{sec:ablation_studies}

\paragraph{Main Ablation on Distillation.}
Figure~\ref{fig:main_ablation} compares success rates and skill-view advantage, defined as the success-rate difference between skill-view and skill-free branches; a sustained negative gap indicates that memories hinder the evolving policy. We ablate forward distillation, cross-view distillation, multi-view rollouts, and distillation. MDSKILL performs best on both benchmarks, with its advantage approaching zero without remaining strongly negative. Removing distillation causes the greatest degradation, while removing multi-view rollouts has the largest impact among variants retaining distillation. The performance of variants retaining multi-view exploration and partial distillation supports the benefits of broader exploration and disentangled knowledge transfer. Crucially, removing forward distillation reduces performance, contrasting with the earlier advantage of reverse-only distillation and showing that skill internalization becomes beneficial within MDSKILL. Cross-view distillation provides further gains through complementary knowledge internalization.

\paragraph{Distillation Stability Analysis.}
Figure~\ref{fig:distillation_stability} compares MDSKILL with
\emph{w/o multi-view rollouts}. Without multi-view rollouts,
forward token overlap fails to improve and the absolute entropy
gap fluctuates sharply, indicating persistent misalignment and
unstable distillation. MDSKILL instead achieves higher overlap
and a smaller, more stable entropy gap, supporting the benefits
of disentangled skill guidance. Cross-view distillation exhibits
the lowest entropy gap and highest overlap, suggesting that
knowledge transfer across complementary views is easier to optimize.

\paragraph{Dynamic Skill Evolution Analysis.}
Figure~\ref{fig:skill_evolution} tracks the mean advantage of top-$K$ skills from each generation cohort. In MDSKILL, high-advantage regions concentrate near the diagonal: new skills address current knowledge gaps, while older skills lose their advantage as their knowledge is internalized. In contrast, UCOB shows weaker diagonal concentration, indicating continued reliance on earlier skills. Figure~\ref{fig:knowledge_evolution} examines skills with substantial advantage ($>0.6$) across 4 levels of generality. MDSKILL shifts from instance-specific knowledge toward cross-task skills and meta-strategies, reducing the final instance-specific share to $26\%$, compared with $42\%$ for UCOB. Although UCOB retains high-advantage skills, this reflects dependence on external memories rather than effective internalization. Together, MDSKILL shows healthy self-evolution, where new skills fill knowledge gaps, internalization reduces their marginal value, and skill evolution shifts toward new and more general knowledge. Additional ablations and analyses are provided in Appendix~\ref{app:additional_ablations}.

\section{Conclusion}
We identify a key obstacle to skill-guided agent self-evolution: noisy and functionally entangled skill memories can provide unstable supervision, hindering knowledge internalization. We propose MDSKILL, a multi-view framework that dynamically selects useful skills, organizes them into complementary functional views, and explores both view-specific and integrated solutions. Outcome-based teacher--student grouping then supports forward, reverse, and cross-view distillation, enabling complementary knowledge to be reliably consolidated into the shared policy. Experiments on three agent benchmarks demonstrate SOTA performance, while ablations and evolution analyses verify more stable distillation and healthier skill evolution. Overall, MDSKILL establishes a self-evolution cycle in which multi-view skill knowledge improves the policy, and the evolving policy continually generates and internalizes better skill memories.

\section*{Reproducibility Statement}
\label{sec:reproducibility}

We provide the information required to reproduce our experiments
throughout the paper. Section~\ref{sec:experiments} describes the
benchmarks, model backbones, baselines, evaluation protocols, and main
results. Appendix~\ref{implementation_details} reports the complete
training, distillation, and skill-retrieval configurations, while
Appendix~\ref{sec:functional_skill_reflection} provides the functional
skill definitions, reflection prompts, and representative examples.

\section*{AI Use Statement}
\label{sec:ai_use}

AI-assisted tools were used during manuscript preparation primarily
to improve grammar, wording, clarity, and LaTeX presentation. They
were not used to design the method, conduct the experiments, or
generate and interpret the reported results. The authors reviewed and
verified all AI-assisted text and take full responsibility for the
final content.

\bibliographystyle{iclr2027_conference}
\bibliography{iclr2027_conference}

\appendix

\section{Related Work}
\label{sec:related_work}

\subsection{Skill Memories for Language Agents}

Retrieval-augmented generation established a general paradigm for conditioning language models on external, non-parametric knowledge~\citep{lewis2020retrieval}. Language agents extend this paradigm from factual documents to interaction experience: Reflexion~\citep{shinn2023reflexion} stores verbal feedback from previous attempts, while Voyager~\citep{wang2023voyager} accumulates executable programs as reusable skills. Recent work further enables agents to construct, select, curate, evolve, or internalize their own skill memories~\citep{xia2026skillrl,mi2026skill,tu2026dynamic,ouyang2026skillos,zhang2026retroagent,lu2026skill0,he2026siri,lin2026skillc,zhang2026co,shi2026skill1,UCOB}. These methods progress from retrieval-time skill augmentation toward evolving external repositories and policy-level knowledge internalization. However, self-generated skills may be noisy, redundant, or functionally entangled; MDSKILL addresses these limitations through dynamic skill selection and multi-view distillation.

\subsection{Agentic Reinforcement Learning}
Proximal Policy Optimization (PPO)~\citep{schulman2017proximal}
stabilizes policy learning through clipped updates, while
GRPO~\citep{shao2024deepseekmath} replaces the learned value
function with relative rewards among grouped rollouts. Building
on these foundations, recent agentic RL methods extend policy
optimization to multi-turn environments involving reasoning,
search, tool use, and environment interaction~\citep{wang2025ragen,
wei2025reinforcing,gigpo,zhu2026gagpo,li2026and,li2026reasoning}.
These methods enable agents to improve from interaction outcomes,
but their supervision remains largely dependent on sparse and
delayed rewards. MDSKILL adopts GiGPO~\citep{gigpo} as its RL backbone because its combination of episode- and step-level optimization is well suited to long-horizon, multi-turn agent interactions.

\subsection{On-Policy Self-Distillation}

On-policy distillation (OPD) provides dense teacher supervision
on student-generated rollouts, reducing off-policy distribution
mismatch~\citep{agarwal2024policy}. On-Policy Self-Distillation
(OPSD)~\citep{opsd} further uses the same policy as teacher and
student under different contexts, with privileged information
provided only to the teacher. This supplies dense token-level
supervision alongside sparse RL rewards without requiring a
separate teacher model. Recent work combines OPSD with agentic RL through skill-conditioned contexts, adaptive supervision, step-level transfer, and bidirectional correction~\citep{yang2026self,wang2026skill,SDAR, zhang2026stepopsd,UCOB}. However, its effectiveness depends on the reliability of retrieved skills. MDSKILL addresses this issue by decomposing skill knowledge into complementary views and performing outcome-guided multi-view distillation.

\begin{algorithm*}[t]
\caption{MDSKILL: Multi-View Distillation Skill.}
\label{alg:mdskill}
\begin{algorithmic}[1]

\Require Policy $\pi_\theta$, reference policy $\pi_{\mathrm{ref}}$,
task distribution $\mathcal{D}$, skill banks
$\mathcal{M}=\{\mathcal{M}_v\}_{v\in\mathcal{V}}$,
retrieval budget $K$, and training steps $T$

\Ensure Trained policy $\pi_\theta$ and evolved skill banks $\mathcal{M}$

\State Define skill views
$\mathcal{V}=\{\mathrm{g},\mathrm{p},\mathrm{c}\}$ and rollout branches
$\mathcal{B}=\{0,\mathrm{g},\mathrm{p},\mathrm{c},\mathrm{cmb}\}$

\For{$r=1,\ldots,T$}
    \State Sample a task batch $\mathcal{X}_r\sim\mathcal{D}$

    \ForAll{$x\in\mathcal{X}_r$}
        \State Initialize a multi-view rollout group for task $x$

        \While{the rollout group is active}
            \State Observe the current anchor state $a$

            \Statex \hspace{\algorithmicindent}
            \stepbox{Step 1: Dynamic multi-view skill retrieval.}

            \ForAll{$v\in\mathcal{V}$}
                \State Retrieve
                $\mathcal{C}_v(a)\gets
                \operatorname{TopK}_{m\in\mathcal{M}_v}
                \operatorname{sim}(a,c_m)$

                \State Sample
                $\tilde q_m^d\sim
                \operatorname{Beta}(\alpha_m^d,\beta_m^d)$,
                $d\in\{\mathrm{rel},\mathrm{adv}\}$

                \State Select
                $\mathcal{S}_{a,v}\gets
                \operatorname{TopK}_{m\in\mathcal{C}_v(a)}^{K_{\mathrm{sel}}}\bigl(\tilde q_m^{\mathrm{rel}}+\tilde q_m^{\mathrm{adv}}\bigr)$
            \EndFor

            \Statex \hspace{\algorithmicindent}
            \stepbox{Step 2: Multi-view rollout exploration.}

            \State Construct skill-free, view-specific, and combined
            contexts $\{P_a^b\}_{b\in\mathcal{B}}$

            \State Advance each branch with
            $\pi_\theta(\cdot\mid P_a^b)$ and store its rollout record
        \EndWhile
    \EndFor

    \State Group rollout records by their shared task and anchor state

    \Statex
    \stepbox{Step 3: Knowledge-gain multi-view distillation.}

    \State Initialize the valid pair set
    $\mathcal{P}\gets\varnothing$

    \ForAll{anchor groups $a\in\mathcal{A}$}
        \State Compute
        $n_a^0 \gets \sum_{\tau\in\mathcal{T}_a^0} y(\tau)$ and
        $n_a^+ \gets \sum_{\tau\in\mathcal{T}_a^+} y(\tau)$
        \State Compute the knowledge-gain gate $\gamma_a$
        using Eq.~\ref{eq:distillation_gates}


        \State Collect
        $\mathcal{P}\gets\mathcal{P}\cup
        \{(\tau_t,\tau_s):
        g_a(\tau_t,\tau_s)=1\}$
    \EndFor

    \State Compute
    $\mathcal{L}_{\mathrm{MD}}\gets
    \operatorname{TokenDistill}(\mathcal{P})$

    \State Update $\theta$ using
    $\mathcal{L}_{\mathrm{RL}}
    +\lambda_{\mathrm{MD}}\mathcal{L}_{\mathrm{MD}}
    +\beta_{\mathrm{KL}}\mathcal{L}_{\mathrm{KL}}$

    \State Update dynamic skill values and view-specific skill banks
\EndFor

\State \Return $\pi_\theta,\mathcal{M}$

\end{algorithmic}
\end{algorithm*}

\section{Implementation Details}
\label{implementation_details}


\subsection{Training.}
Across all three benchmarks, we train MDSKILL for 150 steps with a rollout group size of 8, a training batch size of 16, and a validation batch size of 128. We use an actor learning rate of $10^{-6}$ and response limits of 8192 and 512 tokens, and validate every 25 steps. vLLM rollouts use top-$p=1.0$ and a maximum model length of 10240; validation uses temperature 0.4 and top-$p=1.0$. Each rollout group contains four skill-free trajectories and one skill-guided trajectory for each grounding, planning, control, and combined branch, resulting in a $4{:}1{:}1{:}1{:}1$ allocation. For ALFWorld and WebShop, we use a PPO mini-batch size of 64, a low-variance KL coefficient of 0.01, and an invalid-action penalty coefficient of 0.1; their maximum interaction horizons are 50 and 15 turns, respectively. For Search-QA, we use a PPO mini-batch size of 16, a learning-rate warmup ratio of 0.1, a KL coefficient of 0.001, and an invalid-action penalty coefficient of 0.01. Its maximum search horizon and retained interaction history are both 4 turns. All experiments use eight NVIDIA A100 GPUs. The code is available at the \href{https://anonymous.4open.science/r/MDSKILL-CF53}{https://anonymous.4open.science/r/MDSKILL-CF53}.

\subsection{Distillation.}
We use turn-level, response-scoped Top-$K$ distillation with
$K_{\mathrm{tok}}=32$. For each anchor, we retain at most two
teacher--student pairs independently for each of forward, reverse,
and cross-view distillation. All three types share the same objective
with $\lambda_{\mathrm{MD}}=0.2$. We set the token-level
confidence-gate coefficient to 5.0 and assign action tokens a
distillation weight of 2.0. We use the \textbf{strict gate} for Qwen3-1.7B and the \textbf{relaxed gate} for Qwen2.5-3B-Instruct and
Qwen2.5-7B-Instruct. Appendix~D discusses the effect of gate
selection across model scales.

\subsection{Retrieval.}
We organize skill memories into three functional banks corresponding to state grounding, action planning, and execution control. At each encountered anchor state, we first retrieve the 10 most relevant candidates from each bank using the \texttt{all-MiniLM-L6-v2} sentence transformer, with a relevance threshold of 0.4. For each candidate, we independently draw Thompson samples from its Reliability and Advantage posteriors and rank candidates by their equally weighted sum. Each view-specific branch selects the top three skills from its corresponding bank, whereas the combined branch selects the highest-ranked skill from each bank. Both posteriors are initialized with uniform $\operatorname{Beta}(1,1)$ priors and use a temporal-decay half-life of 20 training steps. Within each functional bank, we further distinguish task-level and state-level memories following UCOB~\citep{UCOB}.


\newlength{\skilldynpanelheight}
\newlength{\skilldyncolumnheight}
\setlength{\skilldynpanelheight}{0.20\textwidth}
\setlength{\skilldyncolumnheight}{0.50\textwidth}



\section{Additional Experimental Results}
\label{app:additional_ablations}

This section provides additional analyses of the hyperparameters and
design choices in MDSKILL. Unless otherwise specified, each
experiment changes only the factor under study while retaining the
same backbone, rollout budget, and optimization settings as the main
experiments.

\noindent
\begin{minipage}[t]{0.42\textwidth}
\vspace{0pt}

\paragraph{Ablation on Distillation Configuration.}
Table~\ref{tab:distillation_config_ablation} evaluates the
distillation weight and loss formulation. A moderate coefficient of
$\lambda_{\mathrm{MD}}=0.2$ performs best on both benchmarks, while
removing distillation or assigning it excessive weight substantially
degrades performance. Joint loss also outperforms separate losses.
It avoids fragmented updates across pairs with different response
lengths and token counts, enabling more efficient and stable
optimization. This formulation also preserves a consistent loss scale
across heterogeneous distillation directions.

\end{minipage}
\hfill
\begin{minipage}[t]{0.56\textwidth}
\vspace{0pt}
\centering
\small
\setlength{\tabcolsep}{6pt}
\renewcommand{\arraystretch}{1.08}

\captionof{table}{
Ablations on the distillation weight and loss formulation.
We report the final overall success rate.
}
\label{tab:distillation_config_ablation}

\vspace{2pt}

\begin{tabular}{@{}lcc@{}}
\toprule
Setting
& \textsc{ALFWorld}
& \textsc{WebShop} \\
\midrule
\multicolumn{3}{@{}l}{\textit{Distillation weight}} \\
\hspace{0.8em}$\lambda_{\mathrm{MD}}=0.0$
    & 81.3 & 46.9 \\
\hspace{0.8em}$\lambda_{\mathrm{MD}}=0.1$
    & 88.3 & 77.3 \\
\hspace{0.8em}$\lambda_{\mathrm{MD}}=0.2$
    & \textbf{93.8} & \textbf{82.8} \\
\hspace{0.8em}$\lambda_{\mathrm{MD}}=0.5$
    & 79.7 & 67.2 \\
\midrule
\multicolumn{3}{@{}l}{\textit{Loss formulation}} \\
\hspace{0.8em}Separate losses
    & 83.6 & 65.6 \\
\hspace{0.8em}Joint loss (Ours)
    & \textbf{93.8} & \textbf{82.8} \\
\bottomrule
\end{tabular}

\end{minipage}

\par\vspace{14pt}

\noindent
\begin{minipage}[t]{0.42\textwidth}
\vspace{0pt}

\paragraph{Ablation on Dynamic Skill Retrieval.}
Table~\ref{tab:dynamic_retrieval_ablation} evaluates the components
of dynamic skill retrieval under the same retrieval and training
budgets. The full strategy performs best on both benchmarks, whereas
relying on relevance alone causes the largest degradation. Removing
either value metric, Thompson sampling, or temporal decay also
reduces performance, demonstrating the complementary benefits of
dynamic value estimation, uncertainty-aware exploration, and
adaptation to the evolving policy.

\end{minipage}
\hfill
\begin{minipage}[t]{0.56\textwidth}
\vspace{0pt}
\centering
\footnotesize
\setlength{\tabcolsep}{5pt}
\renewcommand{\arraystretch}{1.08}

\captionof{table}{
Ablation of dynamic skill retrieval.
We report the final overall success rate.
}
\label{tab:dynamic_retrieval_ablation}

\vspace{2pt}

\begin{tabular}{@{}lcc@{}}
\toprule
Retrieval Strategy
& \textsc{ALFWorld}
& \textsc{WebShop} \\
\midrule
w/o Metrics (Relevance Only)
    & 73.4 & 65.6 \\
w/o Reliability
    & 83.6 & 74.2 \\
w/o Advantage
    & 91.4 & 76.6 \\
w/o Thompson Sampling
    & 88.3 & 72.7 \\
w/o Temporal Decay
    & 90.6 & 76.6 \\
Full Dynamic Retrieval (Ours)
    & \textbf{93.8} & \textbf{82.8} \\
\bottomrule
\end{tabular}

\end{minipage}

\par\vspace{14pt}

\begin{figure*}[!t]
\centering

\captionsetup[subfigure]{
    justification=centering,
    singlelinecheck=false,
    skip=3pt
}

\setcounter{subfigure}{0}

\begin{minipage}[t][\skilldyncolumnheight][t]{0.560\textwidth}
    \vspace{0pt}
    \centering

    \begin{subfigure}[t]{\linewidth}
        \centering
        \begin{minipage}[b][\skilldynpanelheight][b]{\linewidth}
            \centering
            \includegraphics[
                width=\linewidth,
                height=\skilldynpanelheight,
                keepaspectratio
            ]{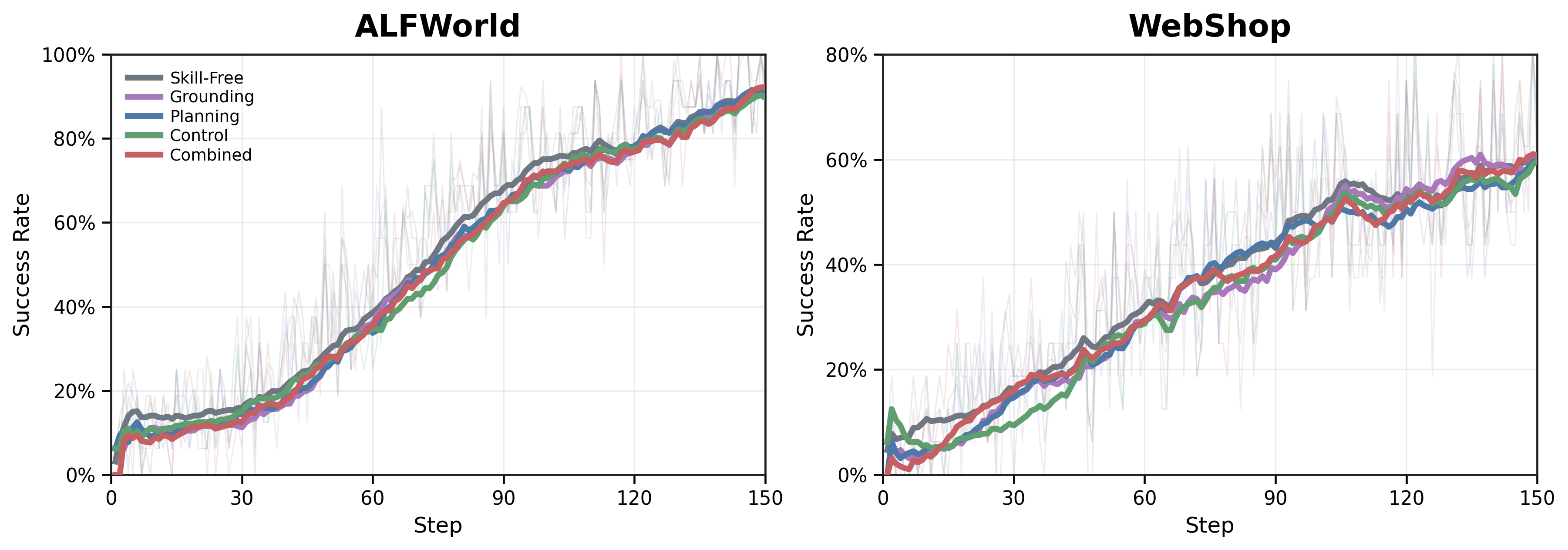}
        \end{minipage}
        \caption{Training success rates of the functional branchs.}
        \label{fig:functional_branch_success}
    \end{subfigure}

    \vfill

    \begin{subfigure}[b]{\linewidth}
        \centering
        \begin{minipage}[b][\skilldynpanelheight][b]{\linewidth}
            \centering
            \includegraphics[
                width=\linewidth,
                height=\skilldynpanelheight,
                keepaspectratio
            ]{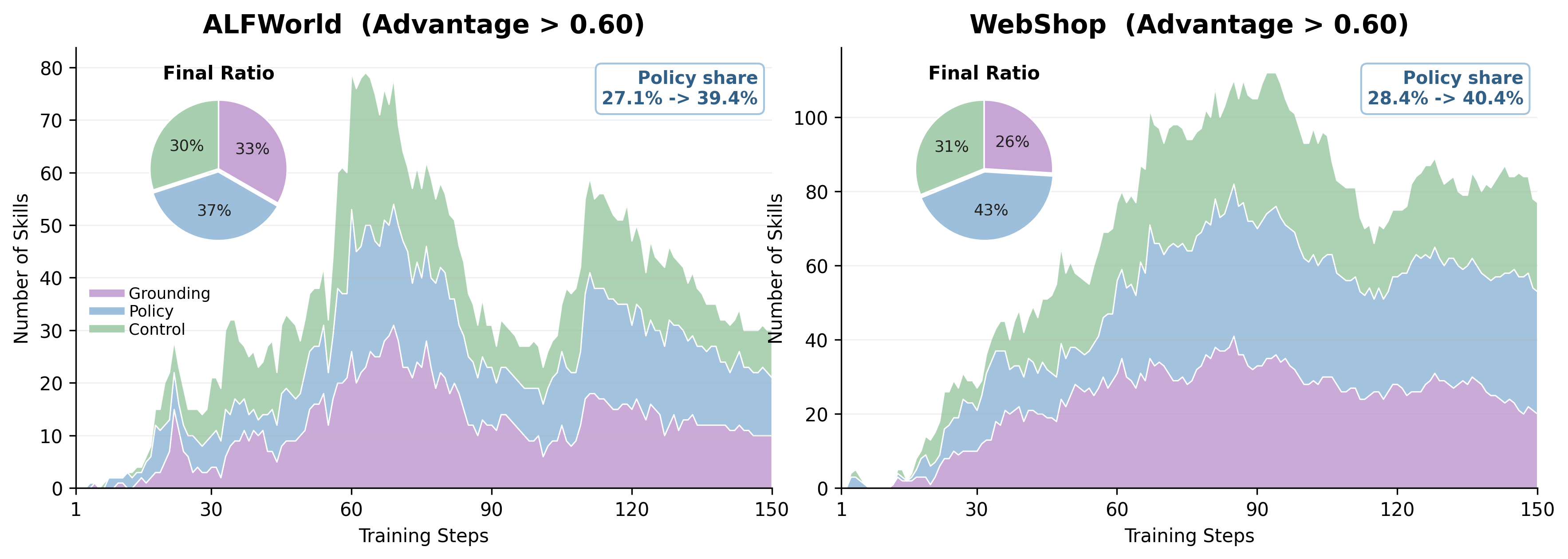}
        \end{minipage}
        \caption{Evolution of high-advantage skills across
        multi-views.}
        \label{fig:functional_skill_advantage}
    \end{subfigure}
\end{minipage}
\hfill
\begin{subfigure}[t]{0.425\textwidth}
    \centering

    \begin{minipage}[t][\skilldyncolumnheight][t]{\linewidth}
        \vspace{0pt}
        \centering

        \begin{minipage}[b][\skilldynpanelheight][b]{\linewidth}
            \centering
            \includegraphics[
                width=\linewidth,
                height=\skilldynpanelheight,
                keepaspectratio
            ]{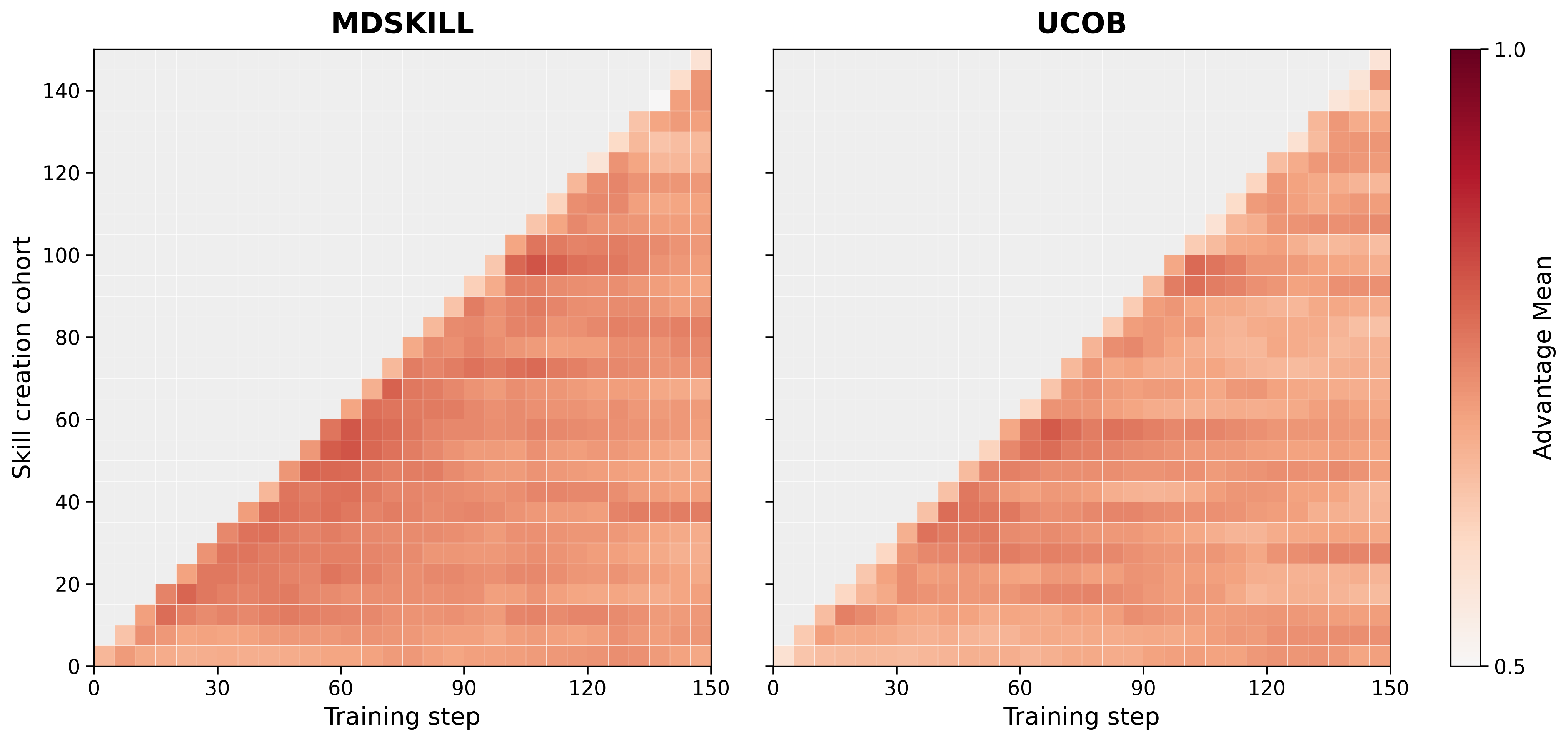}
        \end{minipage}

        \vfill

        \begin{minipage}[b][\skilldynpanelheight][b]{\linewidth}
            \centering
            \includegraphics[
                width=\linewidth,
                height=\skilldynpanelheight,
                keepaspectratio
            ]{images/webshop_horizontal_top10.png}
        \end{minipage}

        \caption{Dynamic skill advantage on ALFWorld (top) and
        WebShop (bottom).}
        \label{fig:skill_advantage_heatmaps}
    \end{minipage}
    
\end{subfigure}

\caption{Additional analyses of functional skill dynamics.
(a) Branch-wise training success rates.
(b) Evolution of skills with substantial Advantage ($>0.6$)
across the functional banks.
(c) Cohort-wise evolution of skill Advantage on ALFWorld and
WebShop.}
\label{fig:additional_skill_dynamics}
\end{figure*}

\paragraph{Functional Skill Dynamics.}
Figure~\ref{fig:additional_skill_dynamics} provides three
complementary views of multi-view skill evolution.
\emph{Branch-wise learning dynamics.}
Figure~\ref{fig:additional_skill_dynamics}(a) shows that all
functional branches improve steadily and remain close to the
skill-free branch on both benchmarks. No individual view exhibits
persistent degradation, indicating that functional decomposition
supports stable multi-view exploration.
\emph{Functional skill composition.}
Figure~\ref{fig:additional_skill_dynamics}(b) tracks skills with
substantial Advantage ($>0.6$). The planning share increases from
$27.1\%$ to $39.4\%$ on ALFWorld and from $28.4\%$ to $40.4\%$ on
WebShop, while grounding and control retain substantial proportions.
This indicates that planning knowledge becomes increasingly valuable
without replacing the complementary contributions of the other
views.
\emph{Cohort-wise skill evolution.}
Figure~\ref{fig:additional_skill_dynamics}(c) shows stronger diagonal
concentration for MDSKILL than for UCOB on both benchmarks: recently
generated skills provide greater Advantage, while older cohorts
gradually lose value as their knowledge is internalized. This pattern
reflects a healthy self-evolution process that continually exposes
new knowledge gaps and encourages the generation of more relevant
skills.

\begin{figure*}[t!]
\centering

\begin{subfigure}[t]{0.485\textwidth}
    \centering
    \includegraphics[
        width=\linewidth,
        trim={0.9cm 0.6cm 0.5cm 1.0cm},
        clip
    ]{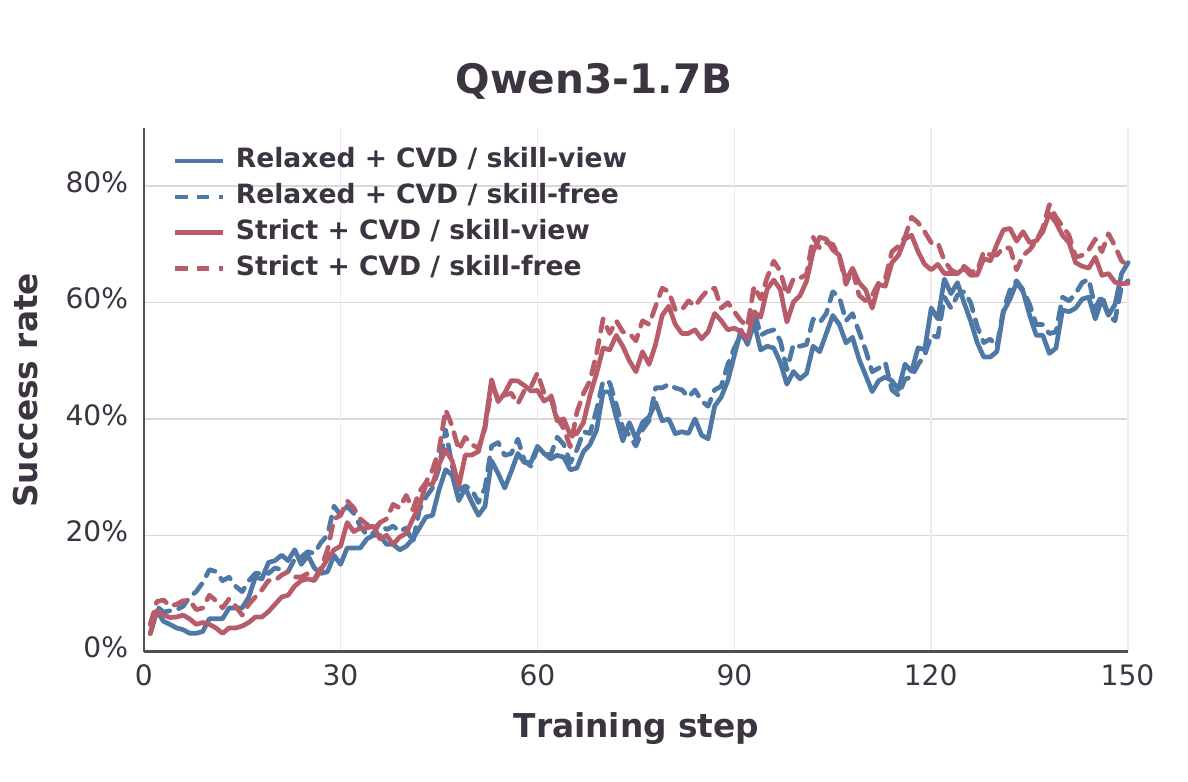}
    \label{fig:distillation_gate_1p7b}
\end{subfigure}
\hfill
\begin{subfigure}[t]{0.485\textwidth}
    \centering
    \includegraphics[
        width=\linewidth,
        trim={0.9cm 0.6cm 0.5cm 1.0cm},
        clip
    ]{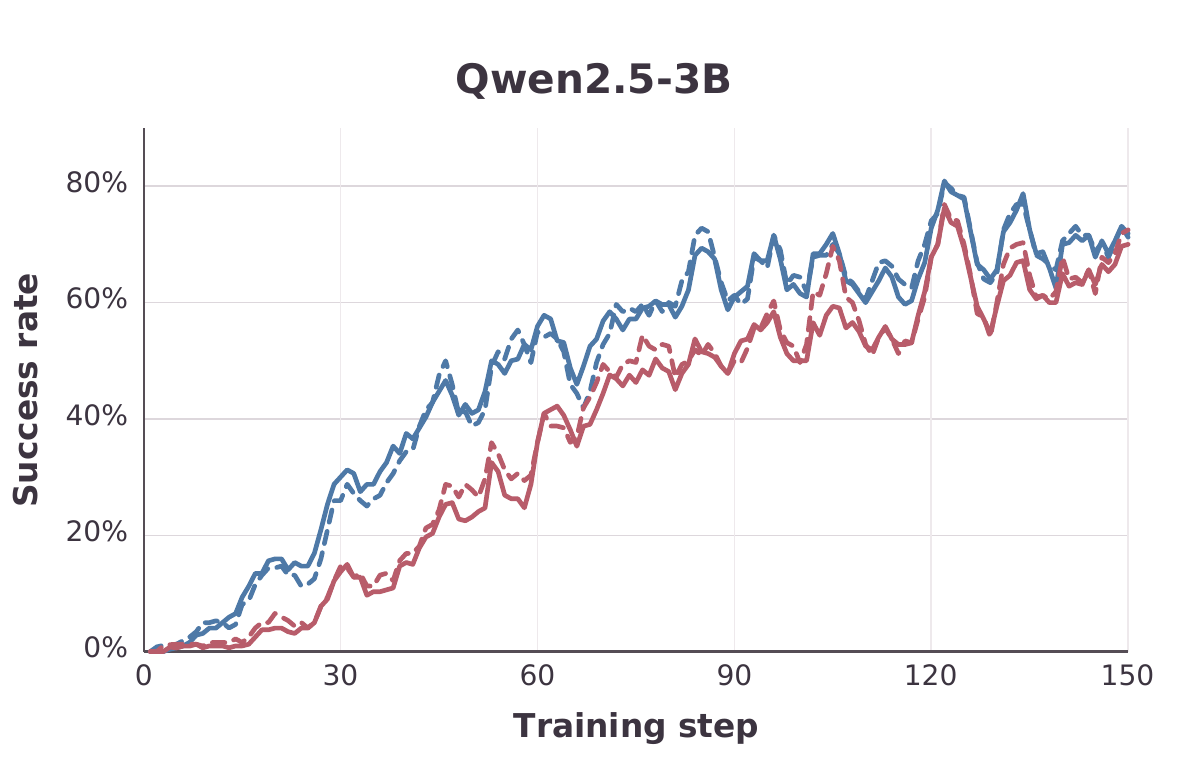}
    \label{fig:distillation_gate_3b}
\end{subfigure}

\caption{Training dynamics under strict and relaxed distillation
gates on WebShop. Solid and dashed lines denote the skill-view and
skill-free branches, respectively.}
\label{fig:distillation_gate_comparison}
\end{figure*}

\section{Discussion on Distillation Gate}
\label{sec:distillation_gate_discussion}

Figure~\ref{fig:distillation_gate_comparison} compares the strict and
relaxed distillation gates across different model scales. For
Qwen2.5-3B, the relaxed gate achieves consistently higher success rates through most of training, and the 7B model exhibits the same trend. In contrast, Qwen3-1.7B benefits more from the strict gate, particularly during the later training stages.

This scale-dependent reversal suggests that the gate reflects the
reliability required of skill-view supervision. Smaller models possess
less task knowledge and tend to generate lower-quality skills, making
skill-view trajectories more likely to provide misleading supervision.
The strict gate permits forward and cross-view distillation only when
all skill-free trajectories fail, filtering ambiguous comparisons
and providing stronger evidence of skill-induced knowledge gain. Larger models generate and utilize more reliable skills; consequently, the relaxed gate can admit more informative teacher--student pairs and enlarge the effective distillation space without introducing excessive noise. Thus, strict filtering is preferable when skill-view supervision is less reliable, whereas relaxed filtering better exploits the stronger
supervision available at larger model scales.

\section{Analysis of Multi-View Distillation}
\label{sec:multiview_analysis}

We provide a local perspective on why multi-view skill decomposition can stabilize distillation. A successful mixed-skill trajectory may draw simultaneously on grounding, planning, and control knowledge, making it difficult to identify which guidance produced useful behavior. Redundant or conflicting guidance may also induce inconsistent token-level gradients. In a view-specific trajectory, the added skill guidance has an explicit functional role. This makes the supervision source easier to interpret, although success alone does not establish that the skill caused the outcome.

Fix an old policy $\pi$ and an anchor state $x$, and condition on successful teacher trajectories with outcome $Y=1$. For this analysis, let $\mathcal{V}_{\mathrm f}$ contain the three functional views: grounding, planning, and control. We use an idealized reference model in which a successful mixed-skill gradient is approximated by first selecting a latent functional source $V\in\mathcal{V}_{\mathrm f}$ and then drawing a stochastic distillation gradient $G$ from the corresponding view-conditioned distribution:
\begin{equation}
p_{\mathrm{mix}}^{\mathrm{ideal}}(G\mid x,Y=1)
=
\sum_{v\in\mathcal{V}_{\mathrm f}}
\alpha_v\,
p(G\mid x,Y=1,V=v),
\label{eq:mixed_skill_gradient_mixture}
\end{equation}
where $\alpha_v=\Pr(V=v\mid x,Y=1)$ in this reference model. This approximation isolates variation in functional composition; it does not model interactions among skills presented together in an actual mixed-skill prompt.

For each view, define its conditional mean gradient and the aggregate mean as
\begin{equation}
\mu_v
=
\mathbb{E}[G\mid x,Y=1,V=v],
\qquad
\bar{\mu}
=
\sum_{v\in\mathcal{V}_{\mathrm f}}\alpha_v\mu_v.
\end{equation}
The gradient variation in the reference mixture decomposes into
\begin{equation}
\begin{aligned}
\sigma_{\mathrm{within}}^2
&=
\sum_{v\in\mathcal{V}_{\mathrm f}}
\alpha_v
\mathbb{E}\!\left[
\|G-\mu_v\|^2
\mid x,Y=1,V=v
\right],\\
\sigma_{\mathrm{between}}^2
&=
\sum_{v\in\mathcal{V}_{\mathrm f}}
\alpha_v
\|\mu_v-\bar{\mu}\|^2.
\end{aligned}
\label{eq:multiview_variance_components}
\end{equation}
The first term captures variation within each view; the second captures differences among their expected gradients. Random changes in view composition introduce both terms into an unstratified batch gradient.

Consider $B$ independent, accepted teacher-gradient instances. Suppose a view-stratified batch contains exactly $B_v=\alpha_vB$ instances from each view. Under the reference model, the unstratified and stratified estimators satisfy
\begin{equation}
\begin{aligned}
\widehat{G}_{\mathrm{mix}}
&=
\frac{1}{B}\sum_{b=1}^{B}G_b,
&
\mathbb{E}\!\left[
\|\widehat{G}_{\mathrm{mix}}-\bar{\mu}\|^2
\right]
&=
\frac{
\sigma_{\mathrm{within}}^2+
\sigma_{\mathrm{between}}^2
}{B},
\\[2pt]
\widehat{G}_{\mathrm{MV}}
&=
\sum_{v\in\mathcal{V}_{\mathrm f}}
\alpha_v
\left(
\frac{1}{B_v}
\sum_{b=1}^{B_v}G_{v,b}
\right),
&
\mathbb{E}\!\left[
\|\widehat{G}_{\mathrm{MV}}-\bar{\mu}\|^2
\right]
&=
\frac{\sigma_{\mathrm{within}}^2}{B}.
\end{aligned}
\label{eq:multiview_gradient_variance}
\end{equation}
Both estimators have mean $\bar{\mu}$ in this model. Fixing the view composition therefore removes its contribution to batch variance:
\begin{equation}
\mathbb{E}\!\left[
\|\widehat{G}_{\mathrm{mix}}-\bar{\mu}\|^2
\right]
-
\mathbb{E}\!\left[
\|\widehat{G}_{\mathrm{MV}}-\bar{\mu}\|^2
\right]
=
\frac{\sigma_{\mathrm{between}}^2}{B}
\geq 0.
\label{eq:multiview_variance_reduction}
\end{equation}
The reduction is strict when the views have different expected gradients. A successful view-specific trajectory also identifies the functional role of its added guidance more clearly than a mixed-skill trajectory, whose success does not isolate the contributions or interactions of its component skills. The variance result concerns batch composition; it does not imply that every individual view-specific trajectory is a better teacher.

To relate variance to optimization, consider a locally fixed, $L$-smooth distillation objective with
$\bar{\mu}=\nabla\mathcal{L}_{\mathrm{MD}}(\theta)$.
For an unbiased gradient estimator $\widehat{G}$ and learning rate
$0<\eta<2/L$, one update satisfies
\begin{equation}
\begin{aligned}
\mathbb{E}\!\left[
\mathcal{L}_{\mathrm{MD}}
(\theta-\eta\widehat{G})
\right]
\leq\;
\mathcal{L}_{\mathrm{MD}}(\theta)
-
\eta
\left(
1-\frac{L\eta}{2}
\right)
\left\|
\nabla\mathcal{L}_{\mathrm{MD}}(\theta)
\right\|^2
+
\frac{L\eta^2}{2}
\mathbb{E}\!\left[
\left\|
\widehat{G}
-
\nabla\mathcal{L}_{\mathrm{MD}}(\theta)
\right\|^2
\right].
\end{aligned}
\label{eq:multiview_descent}
\end{equation}
Thus, under the reference model's assumptions, view stratification yields a tighter expected descent bound for the same number of accepted gradient instances. In MDSKILL, outcome gating and variable response lengths can change the number of accepted instances from each view, so Equations~\ref{eq:multiview_gradient_variance}--\ref{eq:multiview_descent} are not exact variance identities for every training update. They isolate a mechanism by which explicit views can stabilize supervision. Consistent with this mechanism, Figure~\ref{fig:distillation_stability} shows increasing token overlap and a smaller, more stable absolute entropy gap under multi-view rollouts. These distributional measures do not directly estimate gradient variance.


\definecolor{groundingcolor}{HTML}{8B5CC7}
\definecolor{planningcolor}{HTML}{2387D9}
\definecolor{controlcolor}{HTML}{27A567}

\section{Implementation of Functional Skill Reflection}
\label{sec:functional_skill_reflection}

MDSKILL uses view-specific reflection prompts to generate functionally
distinct skills. Figure~\ref{fig:functional_reflection_prompts}
summarizes their differences and presents examples from the trained
ALFWorld skill banks. Shared prompt components are omitted for
readability, with complete templates provided in our code.

\begin{figure*}[t]
\centering

\begin{tcolorbox}[
    colback=gray!3,
    colframe=gray!45,
    boxrule=0.6pt,
    arc=1.5mm,
    left=2.5mm,
    right=2.5mm,
    top=1.2mm,
    bottom=1.2mm,
    fontupper=\footnotesize,
    before upper={
        \setlength{\parindent}{0pt}\raggedright
    }
]
\textbf{Shared prompt components.}
Each prompt receives the task category, task description, trajectory
outcome, a successful reference trajectory when available, and the
trajectory being evaluated. It returns valid JSON containing
\texttt{\{task\_success, task\_category, title, principle,
when\_to\_apply, evidence\}}.
\end{tcolorbox}

\vspace{3pt}

\begin{tcolorbox}[
    enhanced,
    sidebyside,
    sidebyside align=top,
    lefthand width=0.50\linewidth,
    sidebyside gap=4mm,
    segmentation style={
        draw=groundingcolor!65,
        solid,
        line width=0.7pt
    },
    colback=groundingcolor!4,
    colframe=groundingcolor!65,
    colbacktitle=groundingcolor!14,
    coltitle=black,
    title=\textbf{State-Grounding Reflection},
    fonttitle=\footnotesize\bfseries,
    fontupper=\footnotesize,
    fontlower=\footnotesize,
    boxrule=0.8pt,
    arc=2mm,
    left=2.5mm,
    right=2.5mm,
    top=1.5mm,
    bottom=1.5mm,
    before upper={
        \setlength{\parindent}{0pt}\raggedright
    },
    before lower={
        \setlength{\parindent}{0pt}\raggedright
    }
]
\textbf{View-specific instructions.}

Write one high-level grounding skill that helps a future agent
determine the current world state before acting.

\smallskip
\textbf{Think about:}

1. Which room, receptacle, visible object, inventory item, or object
state mattered?

2. What task-relevant fact was unknown or misinterpreted?

3. What observation, navigation, or inspection action exposed that fact?

\smallskip
\textbf{Functional boundary.}
Focus on identifying the current state, not action sequencing or
error recovery.

\tcblower

\textbf{Skill-bank example.}

\textbf{Title:} Kettle State Observation

\textbf{Principle:} Observe the state of the kettle and determine
whether it is clean, dirty, or not visible in the current scene.

\textbf{When to apply:} When determining the state of a kettle in a
cleaning task, such as whether it needs cleaning or is already clean.

\textbf{Evidence:} The agent failed to recognize that the kettle was
not visible or not in the required state and therefore did not proceed
with the cleaning task.
\end{tcolorbox}

\vspace{3pt}

\begin{tcolorbox}[
    enhanced,
    sidebyside,
    sidebyside align=top,
    lefthand width=0.50\linewidth,
    sidebyside gap=4mm,
    segmentation style={
        draw=planningcolor!65,
        solid,
        line width=0.7pt
    },
    colback=planningcolor!4,
    colframe=planningcolor!65,
    colbacktitle=planningcolor!14,
    coltitle=black,
    title=\textbf{Action-Planning Reflection},
    fonttitle=\footnotesize\bfseries,
    fontupper=\footnotesize,
    fontlower=\footnotesize,
    boxrule=0.8pt,
    arc=2mm,
    left=2.5mm,
    right=2.5mm,
    top=1.5mm,
    bottom=1.5mm,
    before upper={
        \setlength{\parindent}{0pt}\raggedright
    },
    before lower={
        \setlength{\parindent}{0pt}\raggedright
    }
]
\textbf{View-specific instructions.}

Write one high-level planning skill that helps a future agent choose
and order actions that advance a similar task.

\smallskip
\textbf{Think about:}

1. Which subgoal was active: search, open, take, clean, heat, cool,
toggle, or place?

2. Which action advanced that subgoal given the known state and inventory?

3. What reusable ordering of subgoals follows from the comparison?

\smallskip
\textbf{Functional boundary.}
Focus on forward task execution, not state discovery or failure
diagnosis.

\tcblower

\textbf{Skill-bank example.}

\textbf{Title:} Clean Kettle and Place in Stoveburner

\textbf{Principle:} First, locate and examine the kettle to determine
its state and location. Then, clean it using an appropriate method and
place the cleaned kettle in the stoveburner.

\textbf{When to apply:} When a task requires cleaning a kettle and
placing it in a stoveburner, but its location or state has not yet
been identified.

\textbf{Evidence:} The agent repeatedly examined unrelated objects
without locating the kettle and consequently failed to clean and place
it in the stoveburner.
\end{tcolorbox}

\vspace{3pt}

\begin{tcolorbox}[
    enhanced,
    sidebyside,
    sidebyside align=top,
    lefthand width=0.50\linewidth,
    sidebyside gap=4mm,
    segmentation style={
        draw=controlcolor!65,
        solid,
        line width=0.7pt
    },
    colback=controlcolor!4,
    colframe=controlcolor!65,
    colbacktitle=controlcolor!14,
    coltitle=black,
    title=\textbf{Execution-Control Reflection},
    fonttitle=\footnotesize\bfseries,
    fontupper=\footnotesize,
    fontlower=\footnotesize,
    boxrule=0.8pt,
    arc=2mm,
    left=2.5mm,
    right=2.5mm,
    top=1.5mm,
    bottom=1.5mm,
    before upper={
        \setlength{\parindent}{0pt}\raggedright
    },
    before lower={
        \setlength{\parindent}{0pt}\raggedright
    }
]
\textbf{View-specific instructions.}

Write one high-level control skill that helps a future agent detect,
avoid, or repair execution errors.

\smallskip
\textbf{Think about:}

1. Which action precondition, task constraint, or completion condition
was violated or overlooked?

2. Did the agent repeat an ineffective action, act on the wrong object,
or advance before verifying the required state?

3. What should the agent verify, retry differently, or backtrack from?

\smallskip
\textbf{Functional boundary.}
Focus on verification and correction, not ordinary planning or
initial state discovery.

\tcblower

\textbf{Skill-bank example.}

\textbf{Title:} Verify Preconditions Before Cleaning

\textbf{Principle:} Verify that the object is in the correct location
and that a cleaning facility, such as a sinkbasin, is available before
performing the cleaning action.

\textbf{When to apply:} When a cleaning action fails because the object
location or required preconditions have not been confirmed.

\textbf{Evidence:} The agent repeatedly examined incorrect objects and
attempted cleaning without first verifying the kettle's location or
the availability of a valid cleaning facility.
\end{tcolorbox}

\caption{View-specific reflection instructions and actual ALFWorld
skills extracted from the trained skill banks. For readability, only
the instructions that differ across functional views are shown; shared
prompt components are omitted and provided in our code. Candidate
skills are written to their intended banks only after
functional-attribution verification.}
\label{fig:functional_reflection_prompts}
\end{figure*}

\paragraph{Functional Skill Generation and Verification.}
For each trajectory comparison, MDSKILL generates grounding, planning,
and control candidates using their respective prompts. Grounding
captures task-relevant states, planning extracts action or subgoal
ordering, and control focuses on execution monitoring and recovery.
Each candidate then undergoes functional-attribution verification and
is admitted to its intended bank only when the predicted category
matches the requested view. This generate--then--verify process reduces
cross-view contamination and preserves functional separation.

\paragraph{ALFWorld Example.}
Figure~\ref{fig:functional_reflection_prompts} presents three skills
generated from an ALFWorld episode requiring a kettle to be cleaned and
placed in a stoveburner. The grounding skill identifies uncertainty
about the kettle's location and cleanliness; the planning skill orders
the locating, cleaning, and placement subgoals; and the control skill
checks the object location and cleaning preconditions. The example
shows how the same trajectory evidence yields distinct, functionally
targeted skills rather than paraphrases of a single lesson.

\end{document}